\RequirePackage{fix-cm}
\documentclass[twocolumn]{svjour3}          %
\smartqed  %
\usepackage{mathptmx}

\usepackage{epsfig}
\usepackage{graphicx}
\usepackage{amsmath}
\usepackage{amssymb}
\usepackage{epstopdf}
\usepackage{makecell,multirow,diagbox}
\usepackage[pagebackref=true,breaklinks=true,letterpaper=true,colorlinks,bookmarks=false,citecolor=blue]{hyperref}
\usepackage{color}
\usepackage{subfigure}
\usepackage{url}

\usepackage{xspace}

\usepackage[numbers,sort]{natbib}
\usepackage{caption}
\begin{document}

\def\Orderless{Orderless\xspace}
\def\Ours{{OBD}\xspace}

\def\multioriented{multi-orientation\xspace}
\def\multiorientedB{Multi-orientation\xspace}

\title{Exploring the Capacity of an \Orderless Box Discretization Network for Multi-orientation Scene Text Detection}

\author{
Yuliang Liu$ ^{1,2}$ 
\and
Tong He$ ^2$
\and
Hao Chen$ ^2$
\and
Xinyu Wang$ ^2$
\and
Canjie Luo$ ^1$
\and
Shuaitao Zhang$ ^1$
\and
Chunhua Shen$ ^{2,3*}$
\and
Lianwen Jin$ ^{1*}$
}

\institute{
  $ ^1$South China University of Technology, China \\
  $ ^2$The University of Adelaide, Australia \\
  $ ^3$Monash University, Australia\\
  $ ^*$Corresponding authors. 
}

\date{Received: date / Accepted: date}

\maketitle

\begin{abstract}
  
  Multi-orientation scene text detection has recently %
  gained 
  significant research attention. Previous methods directly predict words or text lines, typically by using quadrilateral shapes. However, 
  many of these
  methods neglect the significance of consistent labeling, which is important for maintaining a stable training process, especially when it comprises a large amount of data. 
  Here 
  we solve this problem by proposing a %
  new
  method, Orderless Box Discretization (\Ours), which first discretizes the quadrilateral box into several key edges containing all potential horizontal and vertical positions. To decode accurate vertex positions, a simple yet effective matching procedure is proposed for reconstructing the quadrilateral bounding boxes. Our method 
  solves
  the ambiguity issue, which has a significant impact on the learning process. Extensive ablation studies are conducted to validate the effectiveness of our proposed method quantitatively. More importantly, based on \Ours, we provide a detailed analysis of the impact of a collection of refinements, 
  which may %
  inspire others to build state-of-the-art text detectors. Combining both \Ours and these useful refinements, we achieve state-of-the-art performance on various benchmarks, including ICDAR 2015 and MLT.  
  Our method also won the first place in the text detection task at the recent \emph{ICDAR2019 Robust Reading Challenge for Reading Chinese Text on Signboards}, further demonstrating its 
  superior performance. 
  The code is available at {\color{blue}
  \url{https://git.io/TextDet}}.
\end{abstract}

\section{Introduction}\label{sec:introduction}
Scene text detection in arbitrary orientations has garnered significant attention in computer vision because of its numerous potential applications, including augmented reality and robot navigation. Scene text detection is also the foundation and prerequisite for text recognition, which provides a reliable and straightforward approach to scene understanding. However, this challenge remains largely unsolved because text instances in natural images are often of 
multi-orientation, low-quality representations, having perspective distortions of various sizes and scales. 

\begin{figure}[t]
  \begin{minipage}[c]{0.49\linewidth}
    \centering
    \centerline{\includegraphics[width = 4.0cm, height = 3.5cm]{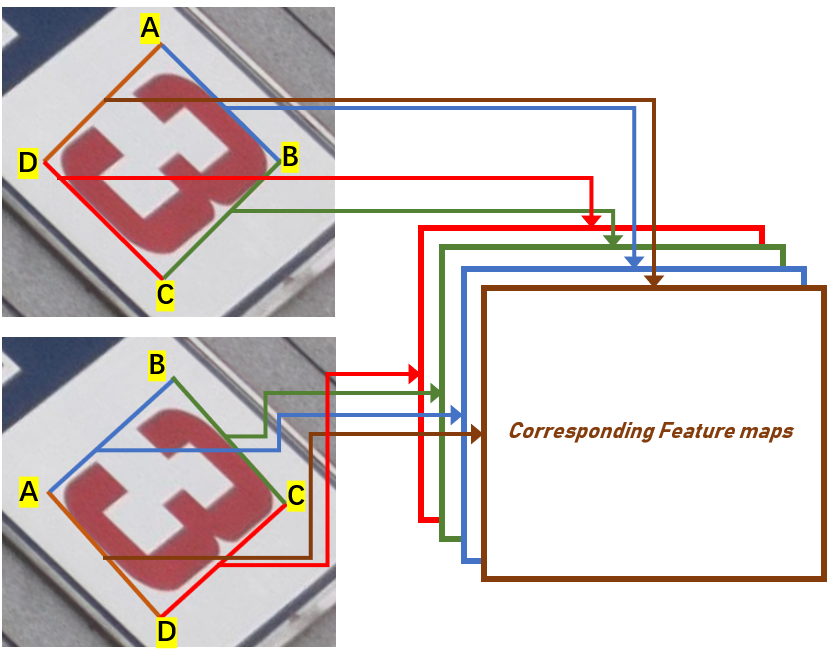}}
    \centerline{\small{\quad (a) Previous regression-based methods.}}\medskip
  \end{minipage}
  \hfill  %
  \begin{minipage}[c]{0.49\linewidth}
    \centering
    \centerline{\includegraphics[width = 4.0cm, height = 3.5cm]{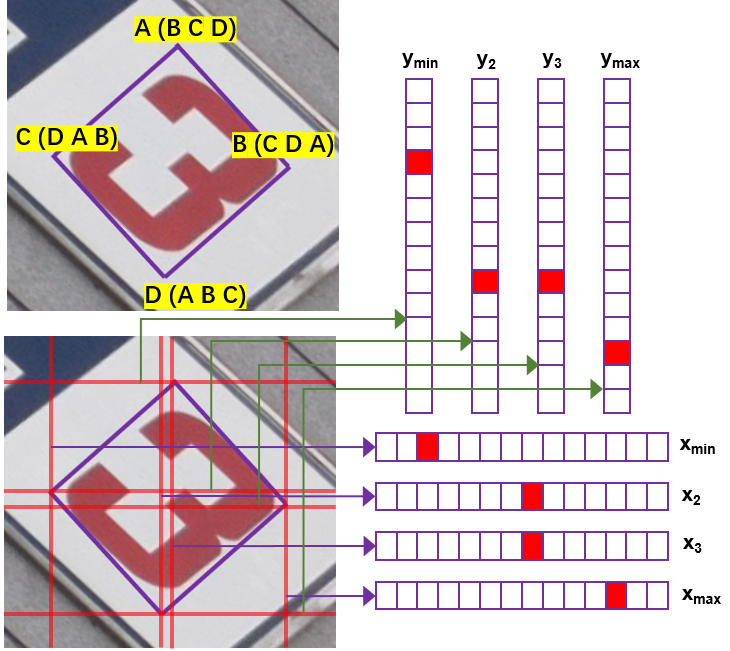}}
    \centerline{\small{(b) Our proposed \Ours. }}\medskip
  \end{minipage}
  
  \caption{Comparison of (a) previous methods and (b) our proposed \Ours. 
  Previous methods directly regress the vertices, which can often be adversely affected 
  by inconsistent labeling of training data, resulting in unstable training and unsatisfactory performances. Our method tackles this problem and removes the ambiguity by discretizing a quadrilateral bounding box that is orderless. 
  }\label{fig:intro_2}
\end{figure}

In the literature, several methods \cite{jaderberg2016reading,neumann2012real,neumann2015real,neumann2015efficient,tian2015text,tian2016detecting} have been developed for solving horizontal scene text detection. However, scene text in the wild is typically presented in a \multioriented form, attracting a few recent studies \cite{zhong2016deeptext,liu2017deep,shi2017detecting,xue2018accurate,xue2018accurate,xie2018scene,liu2019omnidirectional,liao2017textboxes,liao2018textboxes++,liao2018rotation,liu2018fots,he2017single,he2017deep} that can be roughly categorized into two groups: segmentation and regression-based methods. Segmentation-based methods often employ networks, such as fully convolution networks (FCNs) \cite{long2015fully} and Mask R-CNN \cite{he2017mask}. Segmentation-based methods have become 
the
mainstream
approach, because they are sufficiently robust in many complicated scenarios. One limitation is that segmented text instances often require additional post-processing steps. For example, the segmentation results obtained by Mask R-CNN must be fitted into rotated quadrilateral bounding boxes, which necessitates a number of heuristic settings and geometric assumptions.

On the other hand, Regression-based methods \cite{zhu2018sliding,liu2017deep,xue2019msr,liao2018rotation,ma2018arbitrary,liao2018textboxes++,he2018end,zhou2017east,he2017deep} are comparatively simple.
For multi-orientation text, explicitly predicting the vertices obtains the four boundaries of the text instances. Thus, no additional grouping procedure is required. Although these methods can directly predict vertex positions, the significance of regression without facing inconsistent labeling has rarely been discussed. Consider the efficient and accurate scene text (EAST) detector \cite{zhou2017east} method as an example. In EAST, each feature within a text instance is responsible for regressing the corresponding quadrilateral bounding box by predicting four distances to the boundaries and a rotation angle from the viewpoint.
A pre-processing step to assign regression targets is required. As shown in Figure \ref{fig:intro_2}, the regression targets can be altered drastically, even with a minor rotation. Such ambiguities lead to an unstable training process, which considerably degrades the performance. Our experiments indicate that the accuracy of EAST \cite{zhou2017east} 
deteriorates 
sharply (by more than 10\%) when equipped with a random rotation technique for data augmentation, which is supposed to boost the performance.

To address this problem, we propose a novel method, (\textit{i.e.}, \Orderless Box Discretization (\Ours)), which consists of two modules: \textit{Key Edges Detection} and \textit{Matching-Type Learning}. The fundamental idea is to employ invariant representations (\textit{e.g.}, minimum $x$, minimum $y$, maximum $x$, maximum $y$, mean center point, and intersecting point of the diagonals) that are irrelevant to the label sequence to deduce the bounding box coordinates inversely. To simplify the parameterization, the \Ours method first locates all discretized horizontal and vertical edges that contain a vertex. Then, a sequence labeling matching type is learned to determine the best-fit quadrilateral. By avoiding the ambiguity of the training targets, our approach successfully improves performance when a large amount of rotated data is involved.

We complement our method with a few critical technical innovations that further enhance performance. We conduct extensive experiments and ablation studies based on our method to explore the influence of six relevant issues: 
(%
namely, data arrangement, pre-processing, backbone, proposal generation, prediction head, and post-processing) to determine the significance of the various components. 
We
thus 
provide useful tips for designing state-of-the-art text detectors. Leveraging \Ours and these useful refinements, we won first place in the task of Text Line Detection at the \textit{ICDAR2019 Robust Reading Challenge on Reading Chinese Text on Signboards}.

Our main contributions are summarized as follows. 
\begin{itemize}
    \item[1.] Our method addresses the inconsistent labeling issue of regression-based methods, which is of great importance for achieving good detection accuracy.  
    \item[2.] The flexibility of our proposed method allows us to make use of several key refinements that are critical to further boosting accuracy. 
    Our method achieves state-of-the-art performance on various scene text detection benchmarks, including ICDAR2015 \cite{karatzas2015icdar} and MLT \cite{nayef2017icdar2017}. Additionally, our method won the first place in the Text Detection task of the recent \textit{ICDAR2019 Robust Reading Challenge on Reading Chinese Text on Signboard}. Based on the detection results, we integrate 
    advanced 
    recognition models to achieve state-of-the-art results.
    \item[3.] Our method 
    can be generalized to ship detection in aerial images without minimum modification. The significant improvement in terms of the TIoU-Hmean metric further demonstrates the robustness of our approach.
\end{itemize}

\section{Related Work}\label{sec:related_work}

\begin{figure*}[t!]
    \centering
    \subfigure[DMPNet \cite{liu2017deep}.]{\includegraphics[width = 4.5cm, height = 3.2cm]{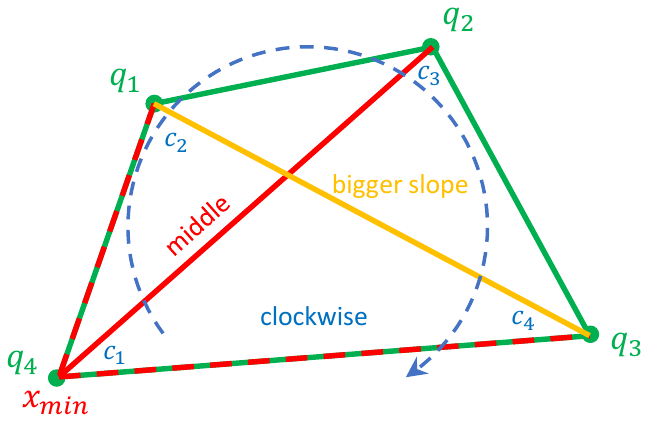}\label{fig:related_work_DMPNET}}
    \hspace{1cm}
    \subfigure[Textboxes++ \cite{liao2018textboxes++}.]{\includegraphics[width = 4.5cm, height = 3.2cm]{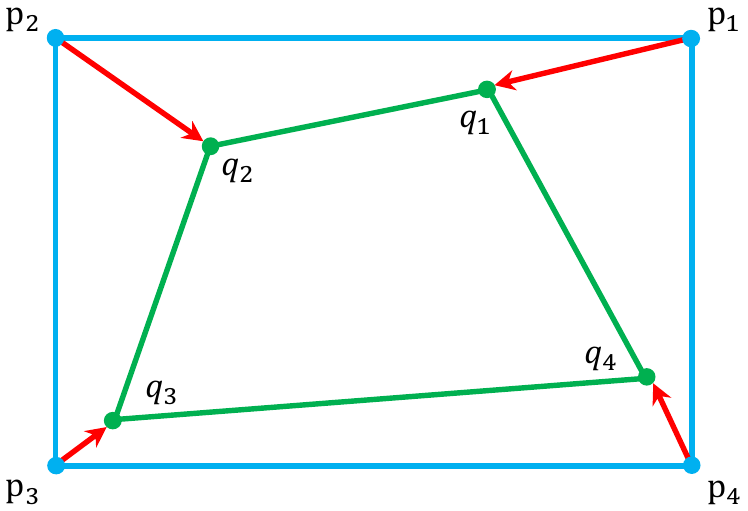}\label{fig:related_work_TEXTBOXES}}
    \hspace{1cm}
    \subfigure[QRN \cite{he2018end}.]{\includegraphics[width = 4.5cm, height = 3.2cm]{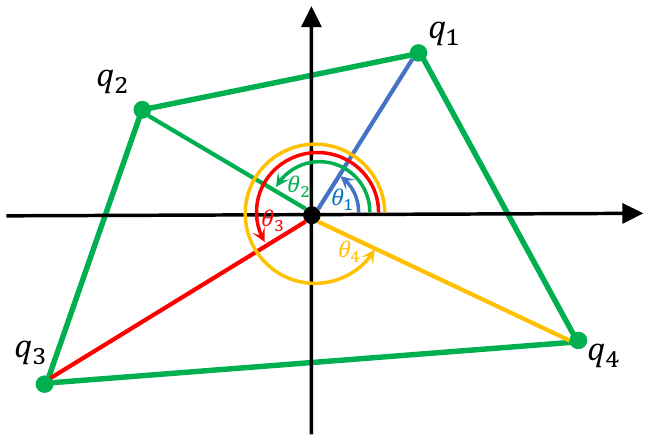}\label{fig:related_work_QRN}}
    \caption{Previous solutions can be negatively affected by the inconsistent labeling issue.}
    \label{fig:related_work}
\end{figure*}

Recently, the emergence of new datasets \cite{ch2019total,liu2019curved,sun2019icdar,chng2019icdar2019} has propelled arbitrarily shaped scene text detection to mainstream research. Multi-orientation scene text detection is one of its most important representations, because multi-orientation scene text comprises most of the text found in real-world visual scenes. The computer-driven detection task remains complex, and there is much room for improvement with regards to decoding multi-orientation text from pictures. Hence, detection benchmarks, such as the MLT \cite{nayef2017icdar2017,nayef2019icdar2019} dataset, are leveraged to refine the process.
However, using quadrilateral bounding boxes can result in some problems for both current segmentation and non-segmentation-based methods.

\paragraph{Segmentation-based.} Segmentation-based methods \cite{zhang2016multi,long2015fully,he2017mask,deng2018pixellink,lyu2018multi,wu2017self,wang2019shape,he2016text} usually require additional steps to group pixels into polygons. 

\paragraph{Non-segmentation-based.} Non-segmentation based methods \cite{zhu2018sliding,xue2019msr,liao2018rotation,ma2018arbitrary,liao2018textboxes++,he2018end,liu2017deep,zhou2017east,he2017deep} can directly learn the exact bounding box for localizing the text instances, but they are easily affected by the label sequence. Usually, such methods use a typical sorting method of the coordinate sequence to alleviate this issue. However, the solutions are not robust because the entire sequence may change even with a small amount of interference. To clarify this, we discuss some of the previous solutions as follows:
\begin{itemize}
    \item Given an annotation having coordinates of four points, a common sorting method of the coordinate sequence to alleviate this issue is to choose the point having the minimum $x$ as the first point, then deciding the rest of the points in a clockwise manner. However, this protocol is not robust. 
    Considering the horizontal rectangle as an example, using this protocol, let us decide that the first point is the top-left point. Thus, the fourth point is the bottom-left point. Suppose
    that
    the bottom-left point moves leftward one pixel (which is possible because of the inconsistent labeling). In that case, the original fourth point becomes the first point, and the whole sequence changes, resulting in very unstable learning. 
    
    \item As shown in Figure~\ref{fig:related_work_DMPNET}, DMPNet \cite{liu2017deep} proposed a protocol that uses the slope to determine the sequence. However, if the diagonal is vertical, leftward, or rightward,  change
    of a pixel
    can result in a completely different sequence.
    
    \item As shown in Figure~\ref{fig:related_work_TEXTBOXES}, given four points, Textboxes++ \cite{liao2018textboxes++} uses the distances between the annotation points and the vertices of the circumscribed horizontal rectangle to {determine} the sequence. However, if $q_1$ and $q_4$ have the same distance to $p_1$, and one pixel rotation can completely change the whole sequence. 
    \item As shown in Figure~\ref{fig:related_work_QRN}, QRN \cite{he2018end} first finds the mean center point of the four given points then constructs a Cartesian coordinate system. Using the positive $x$ axis, QRN ranks the intersection angles of the four points and chooses the point having the minimum angle as the first. However, if the first point is in the positive $x$ axis, one pixel change upward or downward will result in an entirely different sequence.
\end{itemize}
Although these methods \cite{liu2017deep,liao2018textboxes++,he2018end} can alleviate  confusion to some extent, the results can be significantly undermined when using pseudo samples having large degrees of rotation. 

Unlike these methods, our method is the first to directly produce a compact quadrilateral bounding box without complex post-processing. Moreover, it can completely avoid inconsistent labeling issues.

\begin{figure*}[t!]
  \centering
  \includegraphics[width=0.95\textwidth]{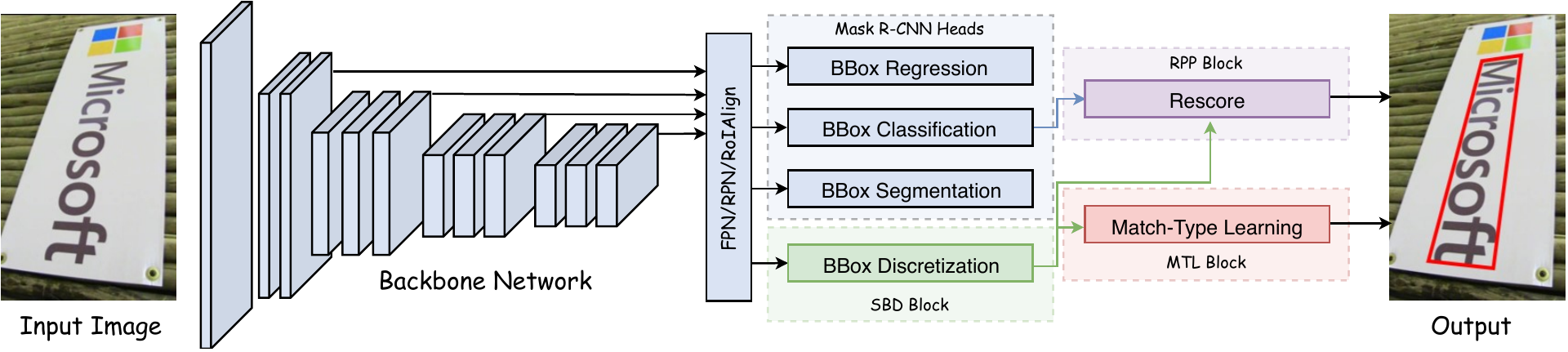}
  \caption{Overview of the proposed detection framework.}
  \label{fig:framework}
\end{figure*}

\section{Our Method}

Our proposed scene text detection system consists of three core components: an \Orderless Box Discretization (\Ours) block, a matching-type learning (MTL) block, and re-scoring and post-processing (RPP) block. Figure~\ref{fig:framework} illustrates the overall pipeline of the proposed framework, and more details are presented in the following sections.

\subsection{ \Orderless Box Discretization}
\label{sec:module}

The purpose of \multioriented scene text detection is to accurately localize the textual content by generating outputs in the form of rectangular or quadrilateral bounding boxes. Compared with rectangular annotations, quadrilateral labels demonstrate an increased capability to cover effective text regions, especially for rotated texts. However, as discussed in Section~\ref{sec:related_work}, simply replacing rectangular bounding boxes with quadrilateral annotations can introduce inconsistency because of the sensitivity of the non-segmentation-based methods to label sequences. As shown in Figure~\ref{fig:intro_2}, the detection model might fail to obtain accurate features for the corresponding points when facing small disturbances. One possible reason behind this is that the neural-network-based regressor for bounding box prediction is essentially a nonlinear continuous function, which means that each input is only mapped to one output. Thus a non-function or a function with a steep gradient cannot be effectively fitted. In our case, a small disturbance may completely change the whole sequence of the vertex and thus a similar input may result in completely different output as well as a steep gradient. Therefore, instead of predicting sequence-sensitive distances or coordinates, an \Ours block is proposed to discretize the quadrilateral box into eight \textit{Key Edges} (KE) comprising order-irrelevant points;
\emph{i.e.}, minimum $x(x_{min})$ and $y(y_{min})$), the second-smallest $x(x_{2})$ and $y(y_{2})$, the second-largest $x(x_{3})$ and $y(y_{3})$, and the maximum $x(x_{max})$ and $y(y_{max})$ (see Figure~\ref{fig:intro_2}). We use x-KEs and y-KEs in the following sections to represent [$x_{min}$, $x_{2}$, $x_{3}$, $x_{max}$] and [$y_{min}$, $y_{2}$, $y_{3}$, $y_{max}$], respectively.

\def\x{$\times$\xspace}

\begin{figure}[b!]
  \centering
  \includegraphics[width=0.47\textwidth]{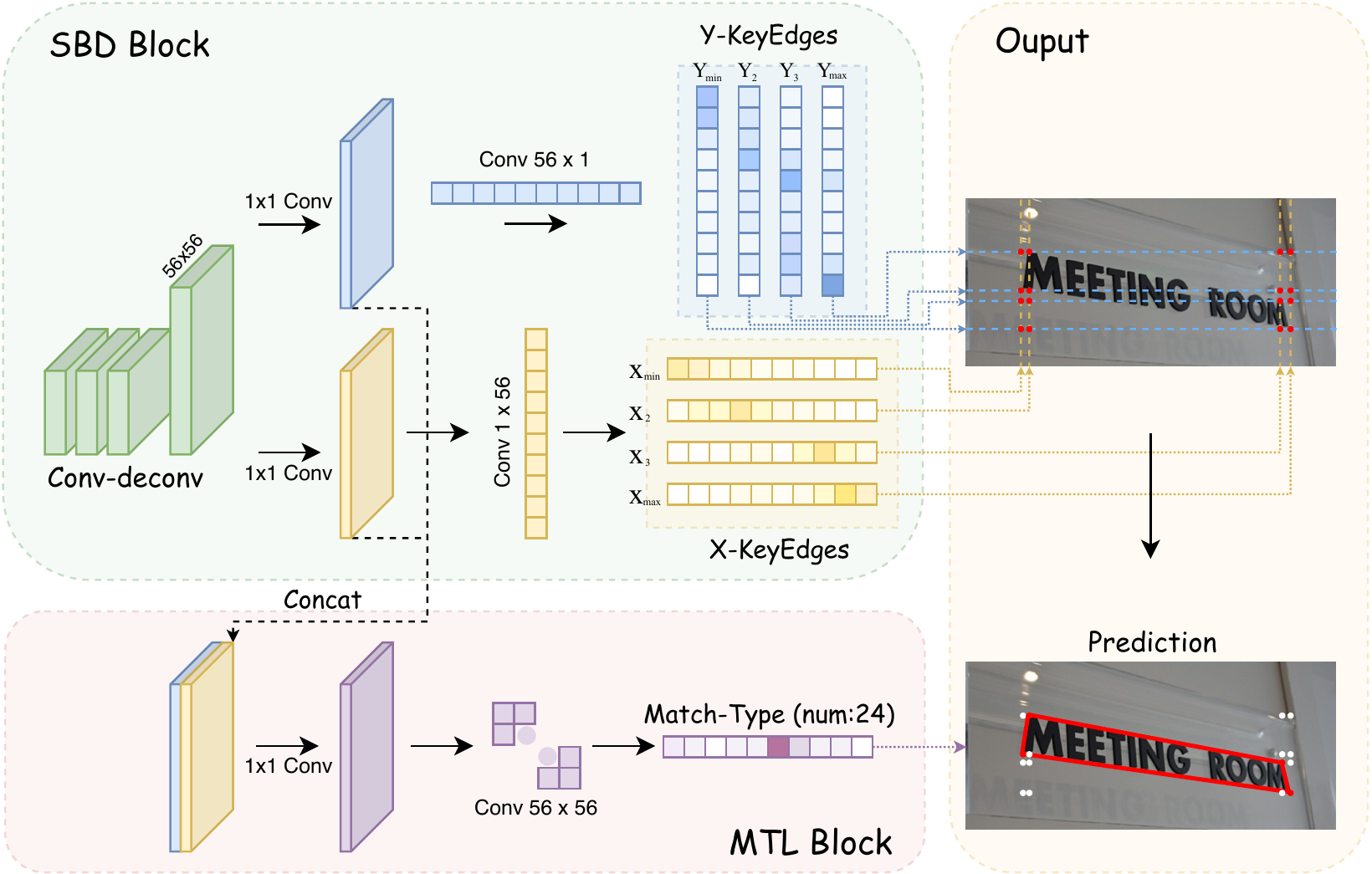}
  \caption{Illustration of the \Ours and MTL blocks.}
  \label{fig:mtl_blocks}
\end{figure}

Specifically, the proposed approach is based on the widely used generic object detection framework, Mask R-CNN \cite{he2017mask}. As shown in Figure~\ref{fig:mtl_blocks}, the proposals processed by RoIAlign are fed into the \Ours block with %
the pooling size of $14\times 14$, where the feature maps are forwarded through four convolutional layers\ with 256 output channels. The output features are then upsampled by a 2\x deconvolutional layer and a 2\x bilinear upscaling layers. Thus, the output size of the feature maps $F_{out}$ is $M \times M$, where $M$ is $56$ in our implementation. 
Furthermore, two convolution kernels shaped as $1 \times M$ and $M \times 1$ with six channels are employed to shrink the horizontal and vertical features for the x-KEs and y-KEs, respectively. Finally, the \Ours model is trained by minimizing the cross-entropy loss $L_{ke}$ over an M-way softmax output, where the corresponding positions of the ground-truth KEs are assigned to each output channel.

\begin{figure}[t!]
	\centering
	\includegraphics[width=8cm]{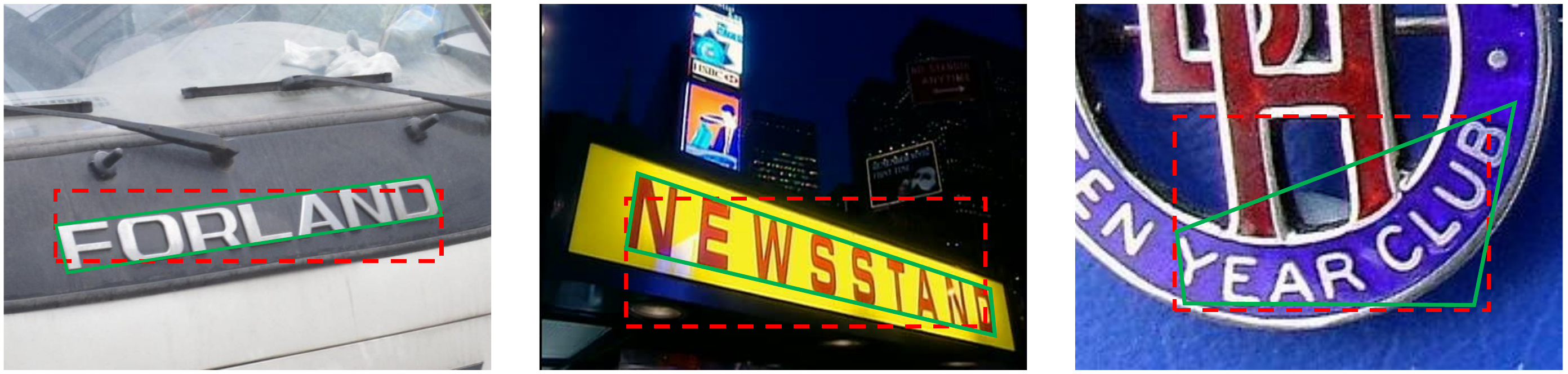}
	\caption{The proposed framework can break the restrictions of the RoIs. The green solid quadrilateral and red dashed rectangular boxes represent the predictions and proposals, respectively.}
	\label{fig:break_restriction}
\end{figure}

In practice, \Ours does not directly learn the x-KEs and y-KEs because of the restriction of the region of interest (RoI). Specifically, the original Mask R-CNN framework limits the prediction inside the RoI areas. Thus, if the regression bounding box is not accurate, the missing pixels outside of the bounding box will not to be restored. To solve this problem, the x-KEs and y-KEs are encoded in the form of ``half lines'' during training. Suppose we have x-KEs, $x^{i} \in [x_{min}, x_{2}, x_{3}, x_{max}]$, and y-KEs, $y^{i} \in  [y_{min}, y_{2}, y_{3}, y_{max}]$. Then, the ``half lines'' are defined as follows:
\begin{equation}
	\begin{array}{c}{\text x^{i}_{half} = \frac{x^{i} + x_{mean}}{2},} \\ \\
	y^{i}_{half} = \frac{y^{i} + y_{mean}}{2},\end{array}
\end{equation}
\noindent where $x_{mean}$ and $y_{mean}$ represent the value of the mean central point of the ground-truth bounding box for the x and y axes, respectively. By employing such a training strategy, the proposed \Ours block can break the RoI restriction (see Figure~\ref{fig:break_restriction}). Thus, it is more likely to produce accurate bounding box because $x_{half}$ and $y_{half}$ fall into the area of the RoIs in most cases, even if the border of the text instance is located outside the RoIs.

Similar to Mask R-CNN, the overall detector is trained in a multi-task manner.  Thus, the loss function comprises four %
terms: 
\begin{equation}
	L = L_{cls} + L_{box} + L_{mask} + L_{ke},
\end{equation}
\noindent where the first three terms, $L_{cls}$, $L_{box}$ and $L_{mask}$, follow the same settings as presented in \cite{he2017mask}.
$L_{ke}$ is the cross-entropy loss, which is used for learning the Key Edges prediction task.
The authors made an interesting observation in which the additional keypoint branch %
can
harm the bounding box detection performance \cite{he2017mask}. However, based on our experiments (see Tables~\ref{tab:ablat} and~\ref{tab:ablat_branch}), the proposed \Ours block is the key component that significantly boosts 
the
detection accuracy. There may be two reasons for this. First, ours is %
different 
from the keypoint detection task, which has to learn $M^{2}$ classes against each other. Thus, the numbers of competitive pixels in the \Ours block is only $M$. Second, for the keypoint detection task, neither one-hot point nor a small circled area can be used to describe the target keypoint accurately, while the KEs produced by \Ours are 
well defined. 
Thus, our method may provide more accurate supervision for training the network.

\begin{figure}[b!]
	\centering
	\subfigure[Correct Matching-type]{\includegraphics[width=4cm,height=3.5cm]{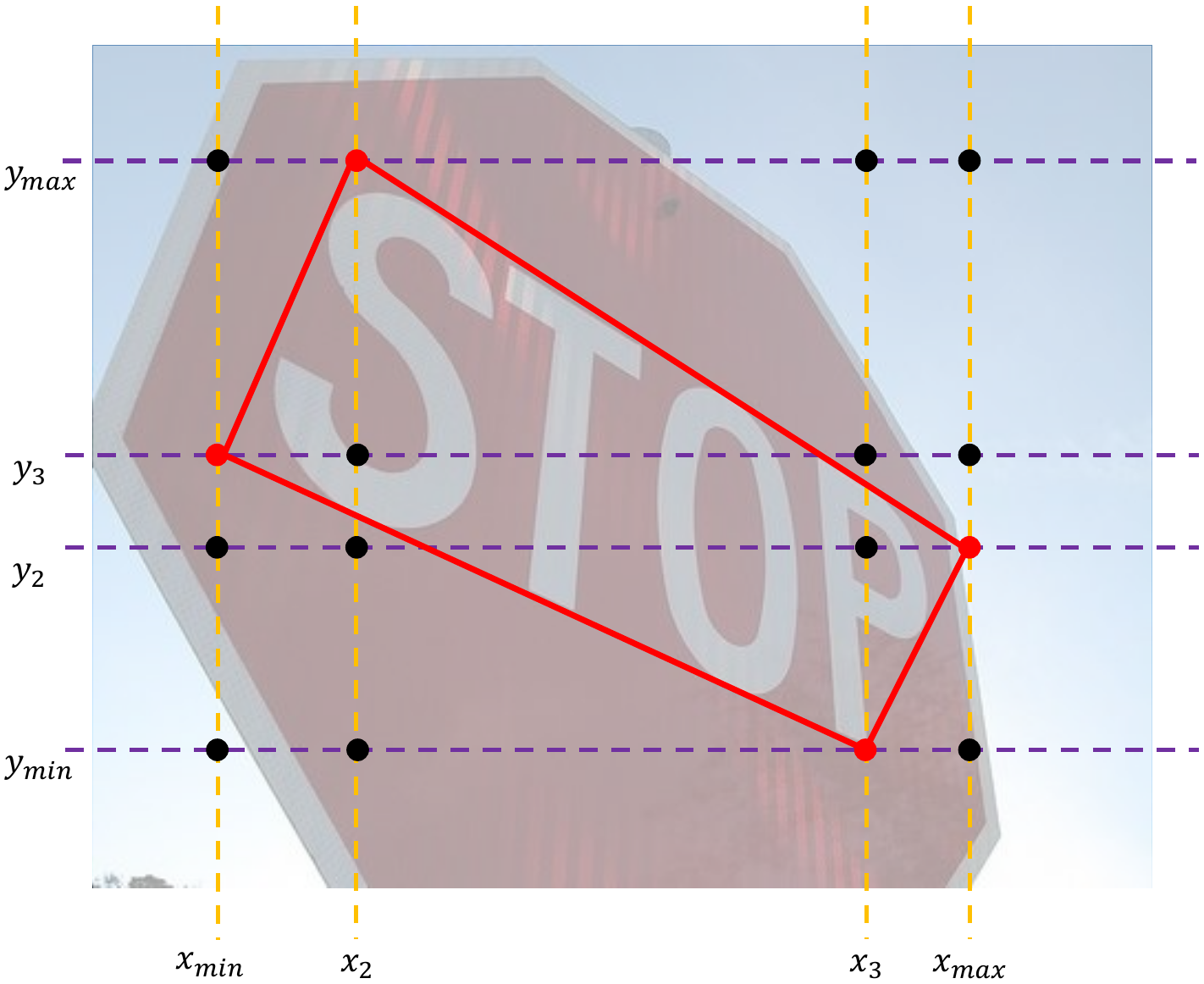}
	\label{fig:match_types_correct}}\hspace{0.05cm}
	\subfigure[Incorrect Matching-types]{\includegraphics[width=4cm,height=3.5cm]{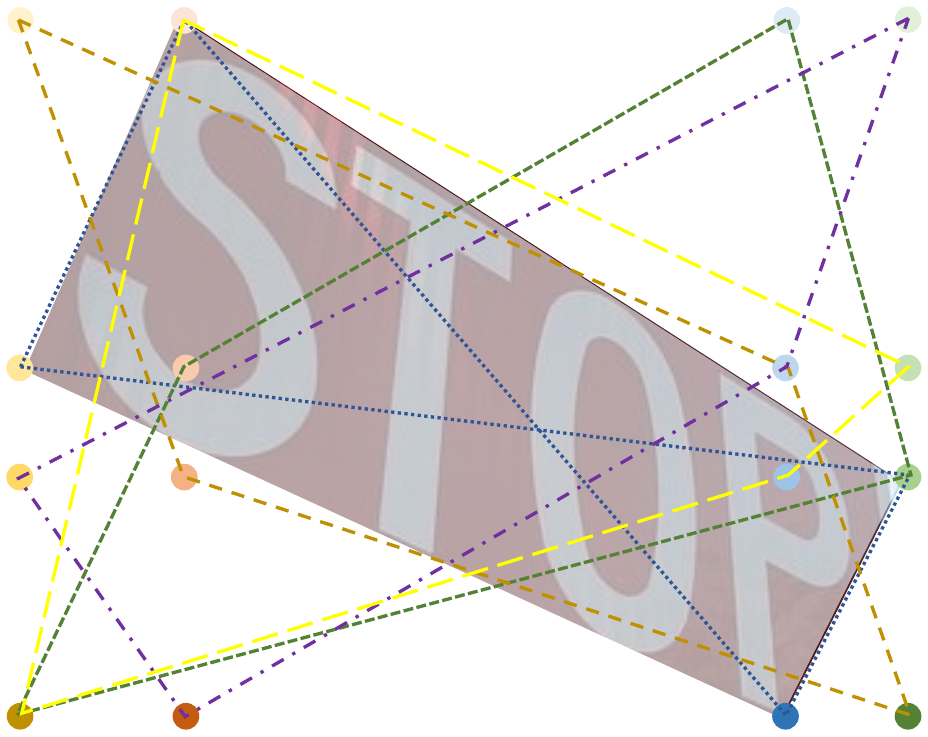}
	\label{fig:match_types_incorrect}}
	\caption{Illustration of different matching types.}
	\label{fig:match_types}
\end{figure}

\subsection{Matching-Type Learning}

It is noteworthy that the \Ours block only learns to predict the numerical values of eight KEs but is unable to predict the connection between the x-KEs and y-KEs. Therefore, we need to design a proper matching procedure to reconstruct the quadrilateral bounding box from the KEs. Otherwise, the incorrect matching type may lead to completely unreasonable results (see Figure~\ref{fig:match_types}).

As described in Section~\ref{sec:module}, there are four x-KEs and four y-KEs outputted by the \Ours block. Each x-KE should match one of the y-KEs to construct a corner point, such as $(x_{min}, y_{min})$, $(x_{2}, y_{max})$, and $(x_{max}, y_{2})$. Then, all four constructed corner points are assembled for the final prediction, giving us the quadrilateral bounding box. It is important to note that different orders of the corners would produce different results. 
Hence, the total number of matching-types between the x-KEs and y-KEs can be simply calculated by $A^{4}_{4} = 24$. For example, the predicted matching-type in Figure~\ref{fig:match_types_correct} is [$(x_{min}, y_{2}), (x_{2}, y_{max}), (x_{3}, y_{min}), (x_{max}, y_{3})$]. Based on this, a simple yet effective MTL module is proposed to learn the connections between x-KEs and y-KEs. Specifically, as shown in Figure~\ref{fig:mtl_blocks}, the feature maps that are used for predicting the x-KEs and y-KEs are used for classifying the matching-types. Specifically, the output feature of the deconvolution layer is connected to a convolutional layer having an $M/2 \times M/2$ kernel size with 24 output channels. Thus, the matching procedure is formed as a 24-category classification task. In our method, the MTL head is trained by minimizing the cross-entropy loss, and the experiments demonstrate that the convergence speed is very fast.

\begin{figure}[t!]
    \centering
    \subfigure[One-peak pattern]{\includegraphics[width=3.8cm, height=3.5cm]{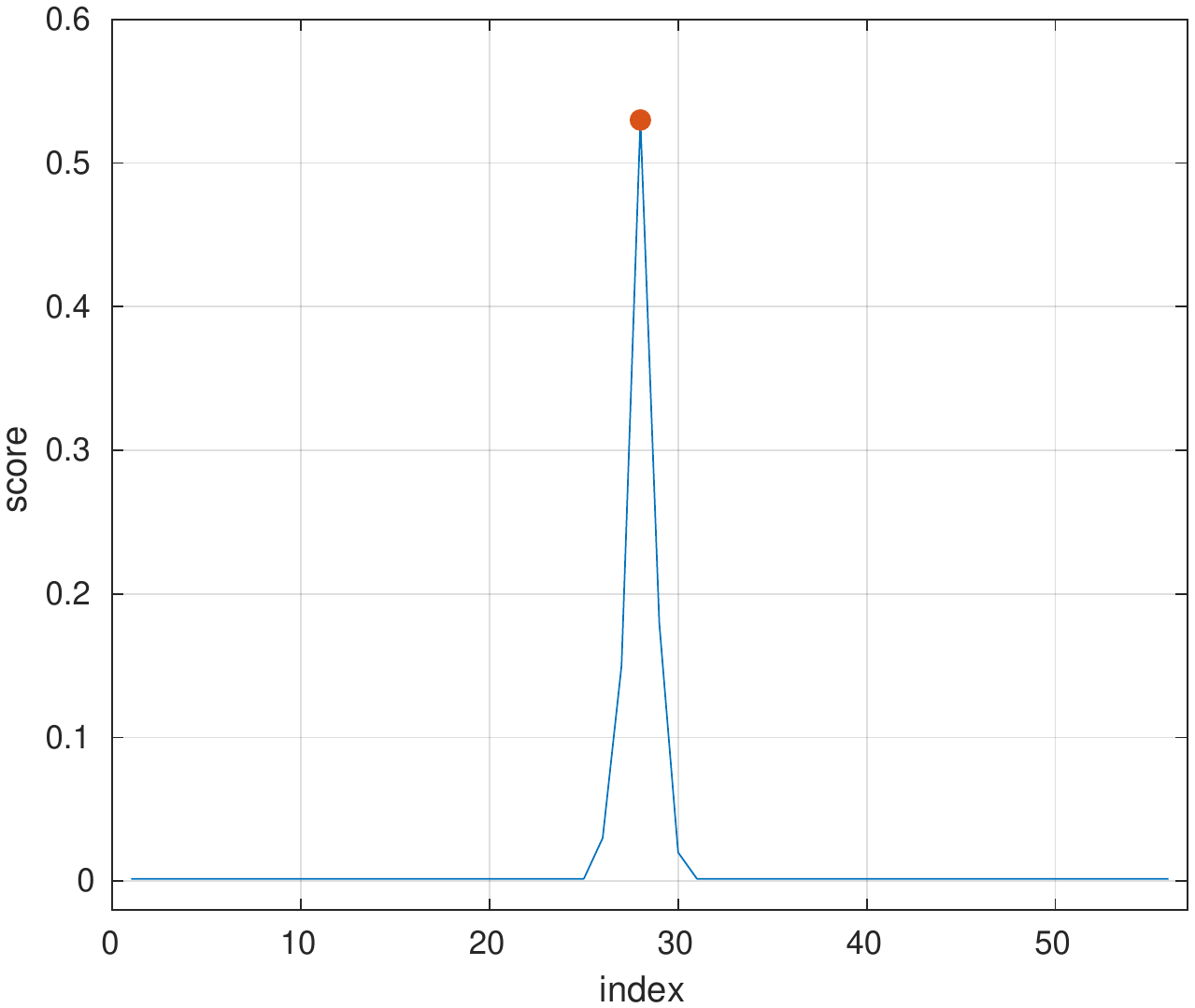}
    \label{fig:KE_score_pattern_norm_1}}
    \hspace{0.05cm}
    \subfigure[Two-peak pattern]{\includegraphics[width=3.8cm, height=3.5cm]{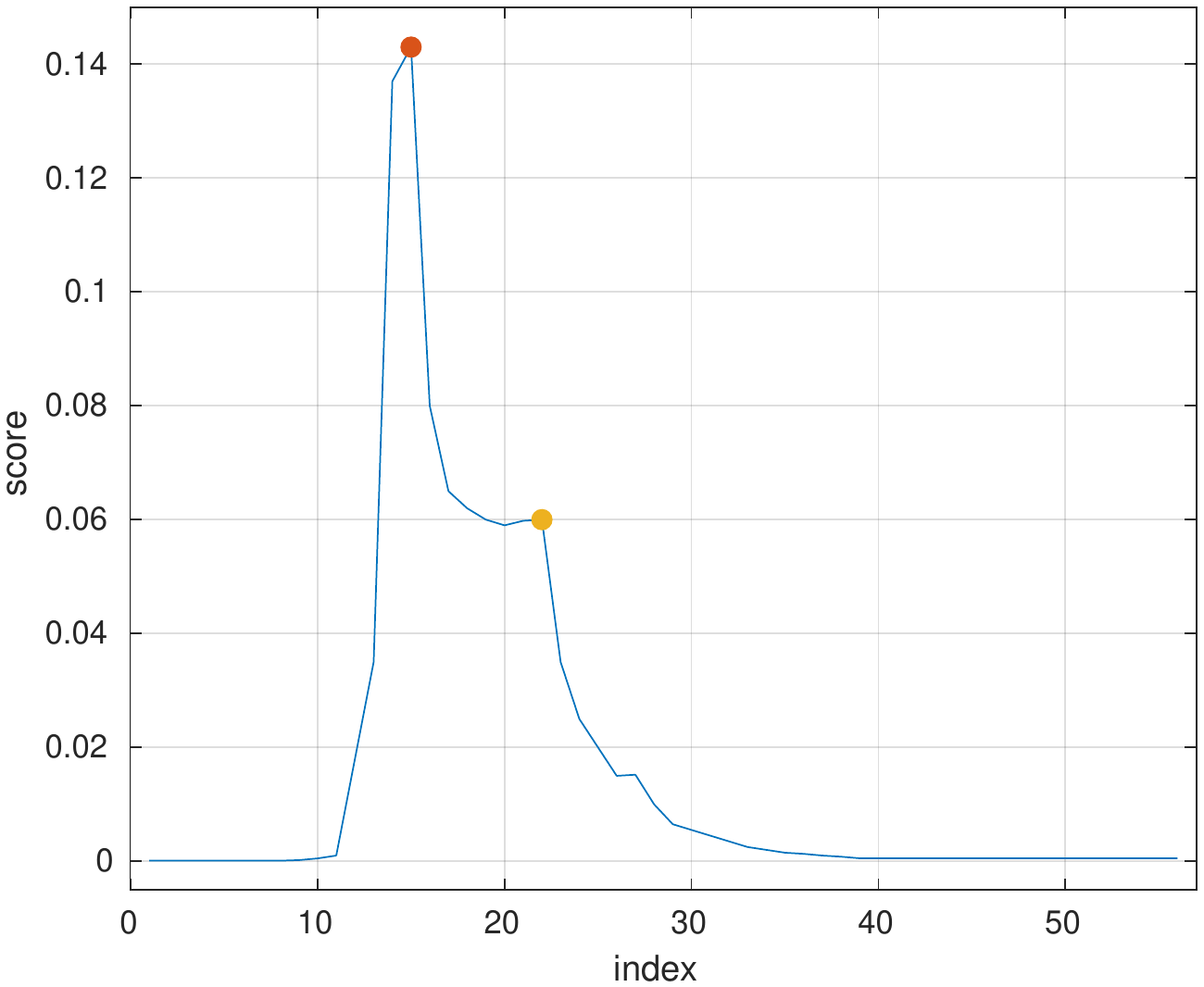}
    \label{fig:KE_score_pattern_abnorm_1}}
    
    \subfigure[Multi-peak pattern]{\includegraphics[width=3.8cm, height=3.5cm]{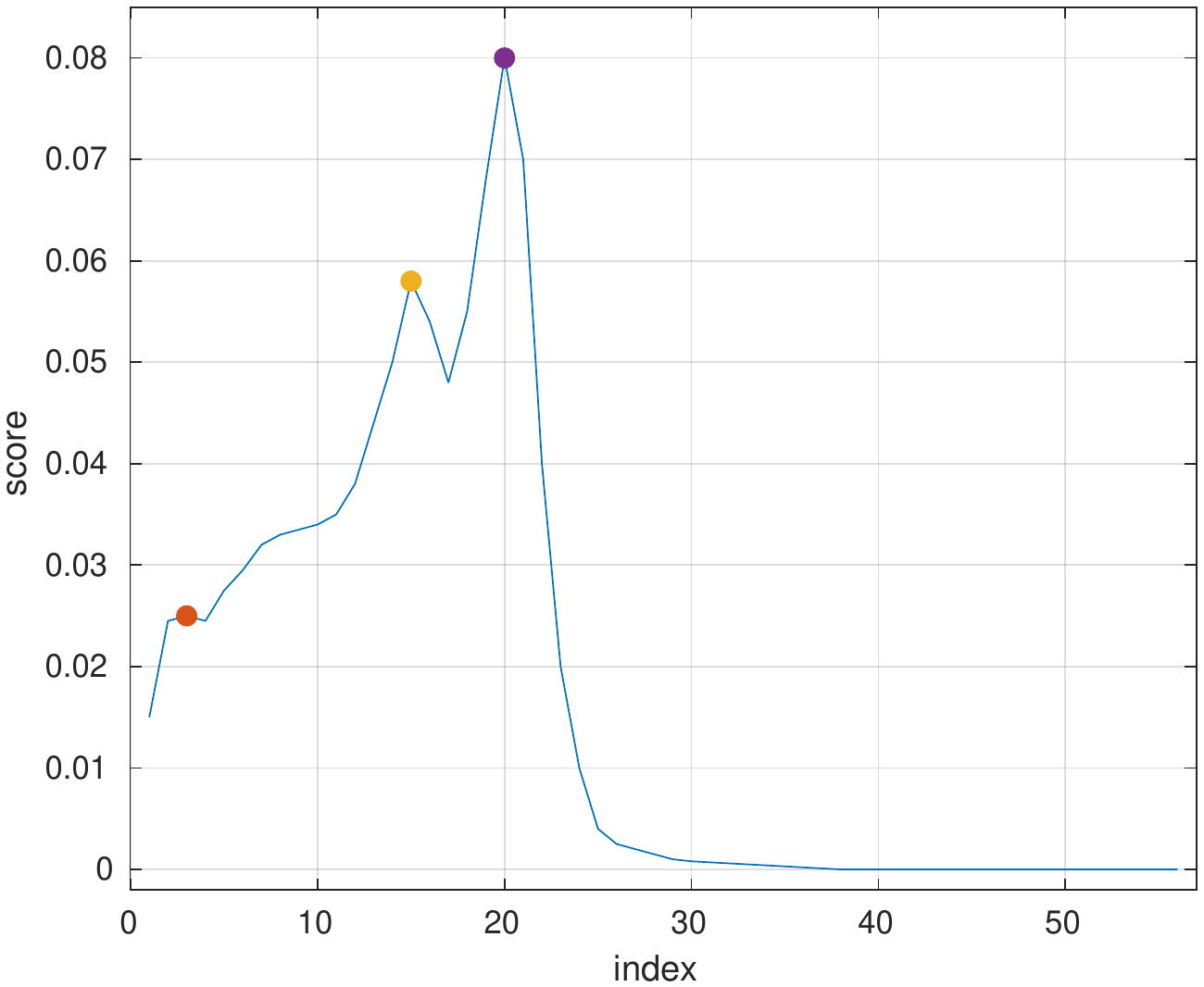}
    \label{fig:KE_score_pattern_abnorm_2}}
    \hspace{0.05cm}
    \subfigure[Multi-peak pattern]{\includegraphics[width=3.8cm, height=3.5cm]{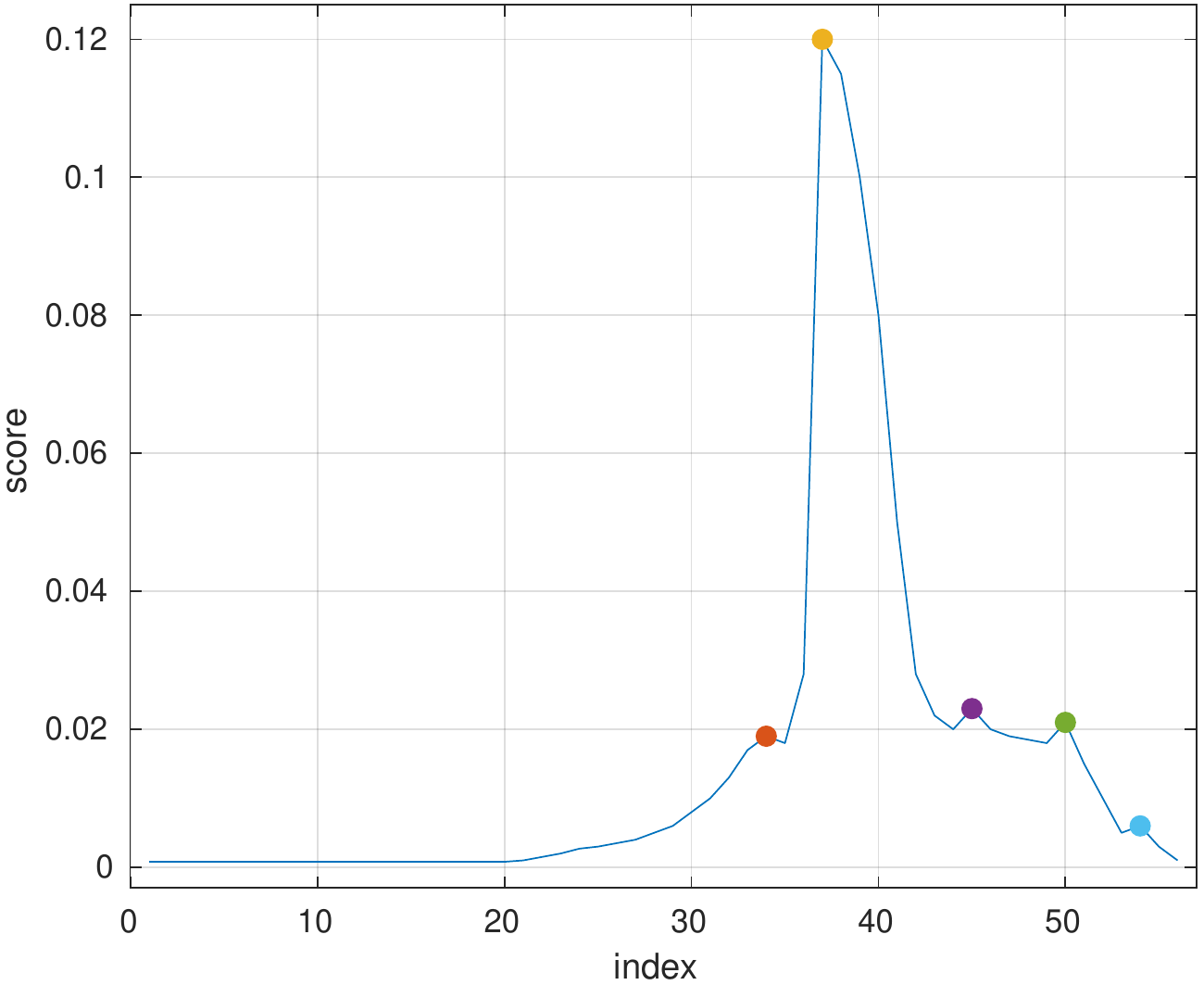}
    \label{fig:KE_score_pattern_abnorm_3}}
    \caption{Different patterns of $S_{   \Ours  }$ outputted by    \Ours   block. (a) is the normal case while (b)(c)(d) are abnormal cases.}
    \label{fig:KE_score_pattern}
\end{figure}

\subsection{Re-scoring and Post-processing}

The fact that the detectors can sometimes output high confidence scores for false positive samples is a long-standing issue in the detection community for both generic objects and text. One possible reason for this may be that the scoring head used in most of the current literature is supervised by the softmax loss, which is designed for classification but not for explicit localization. Moreover, the classification score only considers whether the instance is foreground or background, and it shows less sensitivity to the compactness of the bounding box.

Therefore, a confidence RPP block, is proposed to suppress unreasonable false positives. Specifically, RPP adopts a policy similar to multiple expert systems to reduce the risk of outputting high scores for negative samples. In RPP, an \Ours score $S_{\Ours}$ is first calculated based on eight KEs (four x-KEs and four y-KEs):
\begin{equation}
	S_{   \Ours  } = \frac{1}{K} \sum_{k=1}^{K} \max _{v^{k}} f\left(v^{k}\right),
	\label{eq:score}
\end{equation}
\noindent where $K = 8$ is the number of KEs, $v^{k}$ is the output score vector of the $k^{th}$ KE shown in \eqref{eq:vk}, and $f(v^{k})$ is defined to sum up the peak value, $v_{i}$, and its neighbors.
As shown in Figure~\ref{fig:KE_score_pattern_norm_1}, the distribution of $S_{\Ours}$ demonstrates a one-peak pattern in most cases. Nonetheless, the peak value is still significantly lower than 1. Hence, we sum up four adjacent scores that are near the peak value for each KE score to avoid a confidence score that is too low. 
\begin{equation}
\label{eq:vk}
	v^{k} = [v_{1}, v_{2}, ..., \underbrace{v_{i-2}, v_{i-1}, v_{i}, v_{i+1}, v_{i+2}}_{f(v^{k}) = \sum_{p=max(i-2, 1)}^{P=min(n, i+2)}(v_{p})}, ... v_{n}].
\end{equation}
\noindent It is important to note that the number of adjacent values will be less than four if the peak value is located at the head or tail of the vector. Thus, only the existing neighbors should be counted. Finally, the refined confidence can be obtained by:
\begin{equation}
score=\frac{(2-\gamma) S_{b o x}+\gamma S_{   \Ours  }}{2}, 
\label{eq:rpp_block}
\end{equation}
\noindent where $ 0 \leq \gamma \leq 2$ is the weighting coefficient and $S_{box}$ is the original softmax confidence for the bounding box.
Because both $S_{box}$ and $S_{\Ours}$ are both between [0,1], the value of $score(\Re)$ is also between $ [0,1] $. Counting the $S_{   \Ours  }$ into the final score enables the proposed detector to draw lessons from multiple agents (eight KE scores) while enjoying the benefits of a tightness-aware confidence supervised by the KE prediction task.

\subsection{Discussion}

\begin{figure}[t!]
	\centering
	\includegraphics[width = 8.6cm]{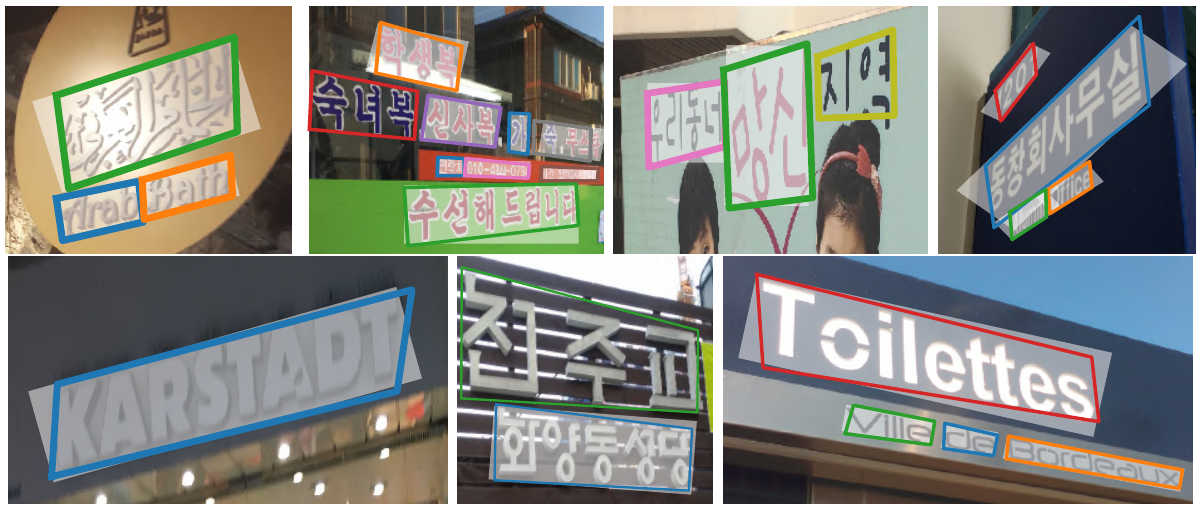} 
	\caption{Compared with the segmentation head, the proposed KE head predicts more compact bounding boxes and shows a higher recall rate for instances that were missed by segmentation. Colored quadrangles are the final detection results, whereas white transparent areas are the mask predictions grouped by the minimum area rectangle.}
	\label{fig:bdn_Ablation }
\end{figure}

It has been proven that the detection performance can be often boosted %
with
the multi-task learning framework. For example, as shown in \cite{he2017mask},  simultaneously training a detection head with an instance segmentation head can significantly improve the detection accuracy. Similarly, a segmentation head is also employed in the proposed \Ours network to predict the area inside the bounding box, which forces the model to regularize pixel-level features to enhance both performance and robustness. However, some issues associated with the segmentation head are highlighted in Figure~\ref{fig:bdn_Ablation }. In (a), the segmentation mask can sometimes produce false positive pixels while the \Ours prediction remains correct. In (b), the segmentation head fails to maintain some positive samples that have been successfully detected by the \Ours block. Therefore, compared with some segmentation-based approaches that directly reconstruct the bounding box by exploiting the segmentation mask, the MTL block can learn geometric constraints to avoid false positives caused by an inaccurate segmentation output. This also reduces the heavy reliance on the segmentation task. Specifically, as shown in Figure~\ref{fig:match_types_incorrect}, the blue dashed line matches an invalid shape that violates the definition of a quadrilateral, because the sides should only have two intersections, at the head and tail. By simply removing these abnormal results, the MTL block can further eliminate some false positives that might cheat the segmentation branch.

Another interesting observation is that the RPP block exhibits a strong capability to suppress false positives, making predictions more reliable. To provide an analysis, we visualize the term $S_{   \Ours  }$, which is used in the RPP block (see Equation~\eqref{eq:rpp_block}). Doing so, we find that there are two typical patterns for the KE scores output by the \Ours block, as shown in Figure~\ref{fig:KE_score_pattern}. Sub-figure (a) shows a one-peak pattern, and sub-figure (b) shows a multi-peak pattern. In normal cases, the KE scores show a regular pattern, in which there is only one peak value in the output vector (see Figure~\ref{fig:KE_score_pattern_norm_1}). However, with hard negative samples, two or more peak values appear (see Figures~\ref{fig:KE_score_pattern_abnorm_1},~\ref{fig:KE_score_pattern_abnorm_2}, and~\ref{fig:KE_score_pattern_abnorm_3}). These multiple peaks share confidence, and the total score is normalized to one. Therefore, based on Equations~\eqref{eq:score} and~\eqref{eq:rpp_block}, the final score will be decreased %
such that 
the proposed model %
is less likely to output
high confidence for those false-positive instances.

Based on our observation, we find that the matching-type prediction could be wrong even if KE is accurate. An example is shown in the bottom instance of the lower-right corner image of Figure \ref{fig:my_label} (b), where $x_{min}$ is mistakenly matched to $y_{min}$. If $x_{min}$ and the second smallest $x$ change their matching $y$ key edge, the detection result can be tighter. Although such a case does not obviously affect both the detection and recognition performance, it is an underlying weakness of the MTL. It is worth mentioning that sometimes the matching type may form an irregular bounding box, \textit{i.e.}, the sides have self-intersection. We find that such cases are very 
rare
and mostly occur with false negatives. For such irregular results, we simply remove  them.

\section{Ablation studies}
\subsection{Implementation details}
Our model is implemented using PyTorch. We %
first evaluate the proposed components of our methods. The initial learning rate is set to 0.01, which is decreased by 10 at 10,000 iterations and 15,000 iterations. The maximum iterations is 20,000 and the image batch size is set to 4. The shorter size of the input image is randomly scaled from 680 to 1000 with the interval of 40, while the maximum size is set to 1480. The weights of KE and matching type learning are set to 0.1 and 0.01, respectively. Flip, random crop, and random rotation are used to improve the generalization ability. Unless %
specified otherwise, 
the re-scoring ratio is kept to be 1.4.

For ablation studies of refinements, each experiment %
uses
a single network that %
is
a variation of our baseline model (first row of Table \ref{tab:Ablation}). Each network %
is
trained on the official ReCTS training set unless specified otherwise. Additionally, because the test scale may significantly influence the final detection result, the testing max size %
is
fixed at 2,000 pixels, and the scale 
is
fixed to 1,400 pixels for strictly fair ablation experiments. The ratio of the flip %
is
also fixed at 0.5, which is the flipping probability for deciding whether to horizontally flip the images for data augmentation. Results are reported on the validation set of ReCTS based on the widely used main performance metric, Hmean. We also report the best confidence threshold that leads to the best performance, which can also reveal some important information.

The number of iterations for training one network %
is set to
80,000 iterations, with a batch size of four images per GPU on four 1080ti GPUs. The final cumulative model %
is
trained %
for 
160 epochs on four V100 GPUs, which %
takes
approximately 6 days. The baseline model %
employs
ResNet-101-FPN as the backbone, which %
is
initialized by a model pretrained on the MLT \cite{nayef2017icdar2017} data. We only %
use
fixed batch normalization for the stem and bottleneck,  i.e., the batch statistics and the affine parameters are fixed. For all prediction heads, we %
do
not use batch normalization. 

\subsection{Ablation studies of the proposed method}
\begin{figure}[htb]
  \centering
  \centerline{\includegraphics[width=8.0cm, height = 4cm]{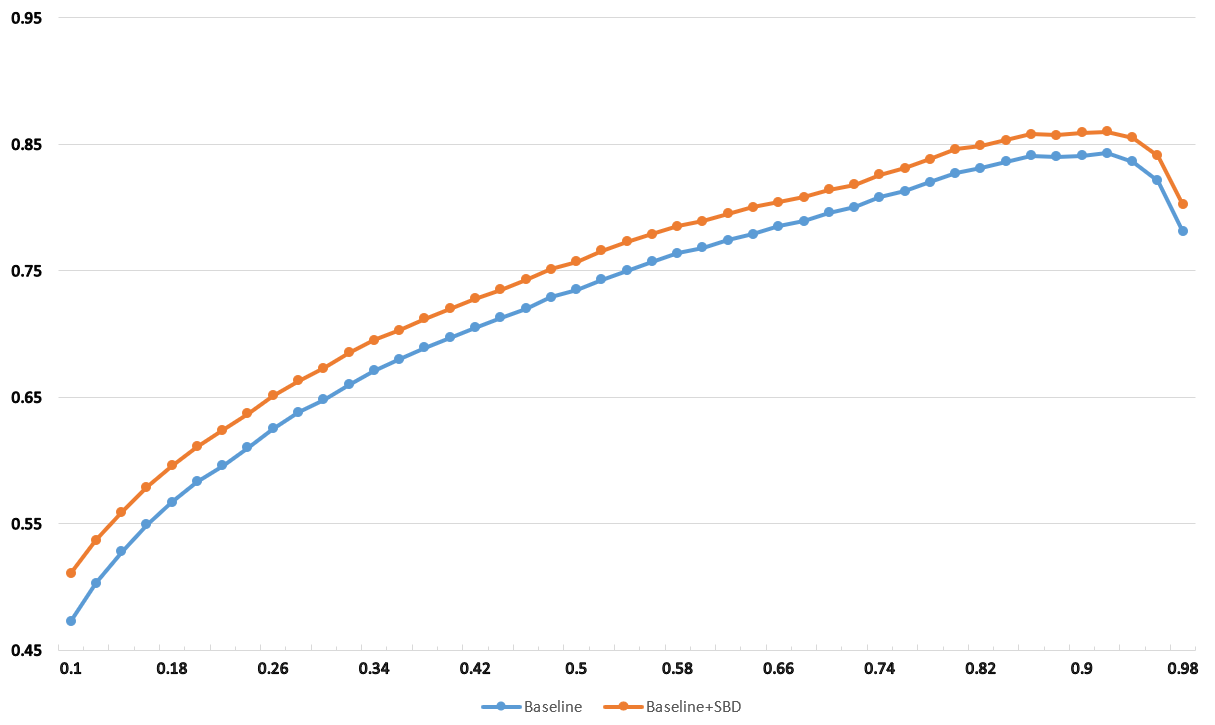}}
  \caption{Ablation study on the ICDAR 2015 benchmark. X-axis represents confidence threshold and Y-axis represents Hmean result. Baseline represents Mask R-CNN. By integrating with proposed \Ours, the detection results can be substantially better than the results of the Mask R-CNN baseline.}\label{fig:ablation}
\end{figure}

In this section, we report ablation studies on the ICDAR 2015 \cite{karatzas2015icdar} dataset,
to validate the effectiveness of each component of our method. 
First, we %
evaluate
the influence of the proposed modules on performance. The results are presented in Table \ref{tab:ablat} and Figure \ref{fig:ablation}. From Table \ref{tab:ablat}, we can see that \Ours and RPP can lead to improvements of 2.4 and 0.6\%, respectively, in terms of Hmean. Additionally, figure \ref{fig:ablation} shows that our method can substantially outperform the baseline Mask R-CNN under different confidence thresholds, further demonstrating its effectiveness.

Furthermore, we conduct experiments by comparing the mask and KE branches (including \Ours and RPP) on the same network. Thus, we test only on one of the branches. We simply use the provided training samples of IC15 without any data augmentation. The results are presented in Table \ref{tab:ablat_branch}, verifying that the proposed modules can effectively improve the scene text detection performance. 

More importantly, we also conduct experiments to verify that introducing ambiguity in the training is harmful to achieving good results.
Specifically, by using the same configuration, we first train Textboxes++ \cite{liao2018textboxes++}, EAST \cite{zhou2017east}, CTD \cite{liu2019curved}, APE \cite{zhu2020adaptive} (the champion method of DOAI2019 competition task1), and the proposed method with the original 1,000 training images of the ICDAR 2015 dataset. Then, we randomly rotate the training images $[0^\circ, 15^\circ, 30^\circ, ..., 360^\circ]$ and randomly 
select
additional 2,000 images from the rotated dataset to fine-tune these models. We also randomly %
select 
additional 2,000 images that are between $[-30^\circ, 30^\circ]$ to evaluate the difference under lower rotation degree.
The results are presented in Table \ref{tab:sequence}. Our method can effectively address the inconsistent labeling issue without drastically degrading the accuracy.
Furthermore, as shown in Table~\ref{tab:ab1_rd}, our proposed method exhibit higher robustness under various degrees of rotation.

Note for the resnet-50 version and the following final competition version of our method, the inference time is 4.5 FPS and 0.83 FPS, respectively. The speed is tested using a single NVIDIA GTX 2080 Ti and the short size of the input image is scaled to 1,000. 

\begin{table}[!t]
\caption{Ablation studies demonstrating the effectiveness of the proposed method. The $\gamma$ of RPP is set to 1.4 (best practice). The results on this table also adopt MLT training data and data augmentation strategies to help improve the final performance.}
\label{tab:ablat}
\centering
\newcommand{\tabincell}[2]{\begin{tabular}{@{}#1@{}}#2\end{tabular}}
\small
\begin{tabular}{c|c|c}
  \hline
  Datasets & Algorithms & Hmean \\
  \hline
  \multirow{3}*{\bf ICDAR2015} & Mask R-CNN baseline & 83.5\% \\
                          \cline{2-3}
                          & Baseline + \Ours  & 85.9\% (\bf  $\uparrow$ 2.4\%) \\
                          \cline{2-3}
                          & Baseline +    \Ours   + RPP & 86.5\% (\bf  $\uparrow$ 3.0\%) \\
  \hline
\end{tabular}
\end{table}

\begin{table}[!t]
\caption{Ablation studies for comparing the mask branch and KE branch. The $\gamma$ of RPP is set to 0.8 (best practice). Compared to Table \ref{tab:ablat}, the results here are all tested in different branches of the same model without any data augmentation. }
\label{tab:ablat_branch}
\centering
\newcommand{\tabincell}[2]{\begin{tabular}{@{}#1@{}}#2\end{tabular}}
\small
\begin{tabular}{c|c|c}
  \hline
  Datasets & Algorithms & Hmean \\
  \hline
  \multirow{3}*{\bf ICDAR2015} & Mask branch & 79.4\% \\
                          \cline{2-3}
                          & KE branch without RPP & 80.4\% (\bf $\uparrow$ 1.0\%) \\
                          \cline{2-3}
                          & KE branch with RPP & 81.0\% (\bf $\uparrow$ 1.6\%) \\
  \hline
\end{tabular}
\end{table}

\begin{table}[!t]
\caption{Comparison on ICDAR 2015 dataset showing different methods' ability of resistant to the inconsistent labeling issue (by adding rotated pseudo samples). TB: Textboxes++. LD: using lower rotation degrees.}
\label{tab:sequence}
\centering
\newcommand{\tabincell}[2]{\begin{tabular}{@{}#1@{}}#2\end{tabular}}
\small
\begin{tabular}{c|c|c|c|c|c}
  \hline
   & TB & East  & CTD & APE  & Ours \\
  \hline
  \tabincell{c}{Hmean \\(baseline)} & 80.1\% & 78.3\% & 74.7\% & 79.4 & 80.4\% \\
  \hline
  \tabincell{c}{Hmean \\(rotation)} & 70.4\% & 64.6\% & 50.1\% & 77.4 & 80.7\% \\
  \hline
  Variance & \bf \textcolor[RGB]{0,160,0}{$\downarrow$ 9.7\%} & \bf \textcolor[RGB]{0,160,0}{$\downarrow$ 13.7\%} & \bf \textcolor[RGB]{0,160,0}{$\downarrow$ 24.6\%} & \bf \textcolor[RGB]{0,160,0}{$\downarrow$ 2.0\%} & \bf \color{red}{$\uparrow$ 0.3\%} \\
  \hline
  \tabincell{c}{Hmean \\(LD)} & 79.5\% & 76.0\% & 68.5\% & 80.1\% & 81.5\% \\
  \hline
  \tabincell{c}{Variance \\ (LD)} & \bf \textcolor[RGB]{0,160,0}{$\downarrow$ 0.6\%} & \bf \textcolor[RGB]{0,160,0}{$\downarrow$ 2.3\%} & \textcolor[RGB]{0,160,0}{$\downarrow$ 6.2\%} & \bf \color{red}{$\uparrow$ 0.7\%} & \bf \color{red}{$\uparrow$ 1.1\%} \\
  \hline 
\end{tabular}
\end{table}

\begin{table}[!t]
\caption{Hmean results under different rotation degrees on ICDAR 2015 dataset. The rotation angle represents the value used for the data augmentation during the training phase.}
\label{tab:ab1_rd}
\centering
\newcommand{\tabincell}[2]{\begin{tabular}{@{}#1@{}}#2\end{tabular}}
\small
\begin{tabular}{c|c|c|c|c}
  \hline
   & 5$^{\circ}$ & 30$^{\circ}$  & 60$^{\circ}$  & 90$^{\circ}$ \\
  \hline
  Ours & $\uparrow$0.9\% & $\uparrow$1.1\% & $\uparrow$1.3\% & $\uparrow$0.3\% \\
  \hline
\end{tabular}
\end{table}

\subsection{Ablation studies of refinements based on our method}
In this section, we provide a detailed analysis of the impact of refinements based on the proposed methods, to evaluate the limits of our method and whether it can be mutually promoted by existing modules. By %
combining
effective refinements, our method achieves first place in the detection task of the ICDAR 2019 Robust Reading Challenge on Reading Chinese Text on Signboards. 

In the following sections, we present an extensive set of experiments that rate our baseline model. Thus, we present results of \Ours having alternative architectures and different strategies with respect to six relevant components for training, including data arrangement, pre-processing, backbone, proposal generation, prediction head, and post-processing.

The objective is to show that the proposed model corresponds to a local optimum in the space of architectures and parameters and to evaluate the sensitivity of the final performance to each design choice. The following discussions follow the structure of Table \ref{tab:Ablation}. Note that the significant breadth and exhaustivity of the following experiments represent more than 3,000 GPU hours of training time.

\subsubsection{Competition Dataset}
\label{subsubsec:rects_cd}
\begin{figure}[!t]
  \centering
  \centerline{\includegraphics[width=8.6cm, height = 6cm]{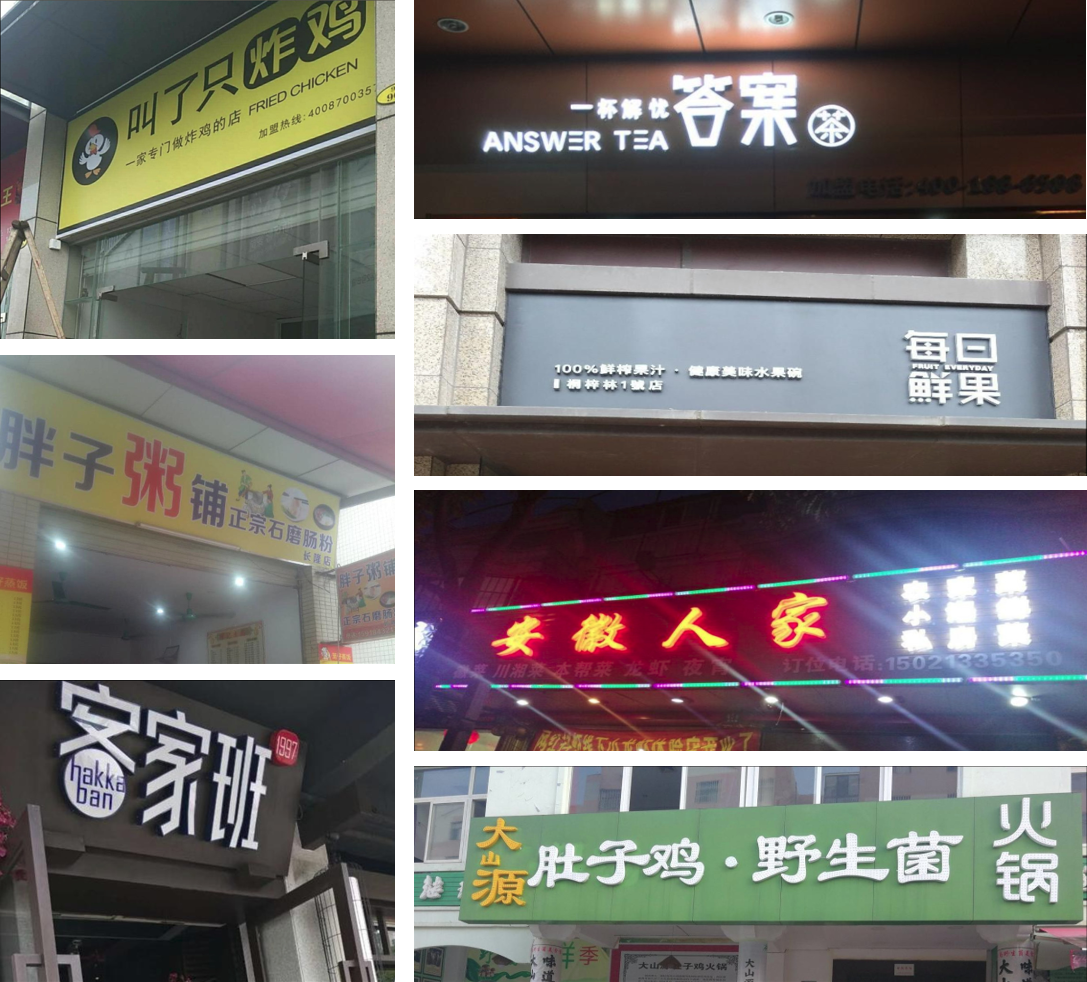}}
  \caption{Example images of the ReCTS. Small, stacked \multioriented, illumination, and annotation ambiguity are the main challenges for this dataset. }\label{fig:ReCTS}
\end{figure}

The competition dataset, Reading Chinese Text on Signboards (ReCTS), is a practical and challenging \multioriented natural scene text dataset containing 25,000 signboard images. A total of 20,000 images are used for the training set, with a total of 166,952 text instances. The remaining 5,000 images are used for the test set. Examples of this dataset are shown in Figure \ref{fig:ReCTS}. The layout and arrangement of Chinese characters in this dataset are clearly different from those in other benchmarks. Because the function of a signboard is to attract a customer base, it is very common to notice their aesthetic appearance. Thus, the Chinese characters can be arranged in any kind of layout with various fonts. Additionally, characters from one word can be in diverse orientations, diverse fonts, or diverse shapes, which complicates the challenge. This dataset provides both text lines and character annotations to inspire new algorithms that can take advantage of the arrangement of characters. To evaluate the function of each component, we split the original training set into 18,000 training images and 2,000 validation images.

\begin{table*}
  \caption{Ablation studies of different refinements based on our method. Each variation is evaluated on the ReCTS validation set. It is worth mentioning that we regard difficult samples as true negatives in the validation because they cannot be recognized and only loosely annotated in the competition dataset. However, in the final ranking, detection box in the difficult region are set to ``do not care'', which can result in a leap improvement.
  We evaluate variations of our baseline model (second row). Every row corresponds to one variation in different part. We train each variation with ResNet-101-FPN and fixed random seeds and equal 80,000 iterations (unless specifying) and report Hmean in the best confident threshold (grid search).}
  \label{tab:Ablation}       %
  \newcommand{\tabincell}[2]{\begin{tabular}{@{}#1@{}}#2\end{tabular}}
  \small
  \centering
  \begin{tabular}{lccccc}
  \hline\noalign{\smallskip}
  Methods & Best threshold & Recall (\%) & Precision (\%) & Hmean (\%) & $\Delta$Hmean  \\
  \hline\noalign{\smallskip}
  \tabincell{c}{\bf Baseline model (based on \Ours \cite{liu2019omnidirectional}) \\ with mlt pretrained model\\ with flip (0.5) \\ test scale: min size: 1400; max size: 2000.} & \bf 0.91 & \bf 78.1 & \bf 80.1 & \bf 79.1 & \bf -  \\
  \noalign{\smallskip}\hline
  {\bf Data arrangement} \\
  \hspace{5mm} With data cleaning & 0.93 & 77.7 & 80.3 & 79.0 & \bf \textcolor[RGB]{0,160,0}{$\downarrow$ 0.1} \\
  \hspace{5mm} With only mlt pretrained data (100k iters)  & 0.97 & 53.4 & 56.1 & 54.7 & \bf \textcolor[RGB]{0,160,0}{$\downarrow$ 24.4} \\
  \hspace{5mm} With only 60k pretrained  data (200k iters) & 0.81 & 50.8 & 61.0 & 55.5 & \bf \textcolor[RGB]{0,160,0}{$\downarrow$ 23.6} \\
  \hspace{5mm} With defect data  & 0.91 & 75.8 & 72.5 & 74.1 & \bf \textcolor[RGB]{0,160,0}{$\downarrow$ 5.0} \\
  \hspace{5mm} Without MLT data pretrain & 0.85 & 75.5 & 81.9 & 78.6 & \bf \textcolor[RGB]{0,160,0}{$\downarrow$ 0.5} \\
  \hspace{5mm} With 60k pretrained model  & 0.91 & 78.8 & 81.9 & 80.3 & \bf \color{red}{$\uparrow$ 1.2} \\
  \noalign{\smallskip}\hline
  {\bf Pre-processing} \\
  \hspace{5mm} With random crop (best ratio)  & 0.91 & 78.4 & 83.7 & 81.0 & \bf \color{red}{$\uparrow$ 1.9} \\
  \hspace{5mm} With random rotate (best ratio)  & 0.91 & 77.6 & 81.8 & 79.7 & \bf \color{red}{$\uparrow$ 0.6} \\
  \hspace{5mm} With color jittering  & 0.91 & 76.4 & 82.5 & 79.3 & \bf \color{red}{$\uparrow$ 0.2} \\
  \hspace{5mm} \tabincell{c}{With medium random scale training \\ \hspace{5mm} ori: (560,600,...,920,) max: 1300 \\ \hspace{5mm} to: (680,720,...,1120,) max: 1800 } & 0.89 & 80.3 & 82.2 & 81.3 & \bf \color{red}{$\uparrow$ 2.2} \\
  \hspace{2mm} \tabincell{c}{With large random scale training \\ \hspace{8mm} ori: (560,600,...,920,) max: 1300 \\ \hspace{8mm} to: (800,840,...,1400,) max: 2560 } & 0.89 & 80.2 & 83.6 & 81.9 & \bf \color{red}{$\uparrow$ 2.8} \\
  \noalign{\smallskip}\hline
  {\bf Backbone} \\
  \hspace{4mm} \tabincell{c}{With ResNext-152-32x8d-FPN-IN5k \\ (using detectron  pretrained model) v1} & 0.91 & 79.4 & 84.0 & 81.6 &  \bf \color{red}{$\uparrow$ 2.5}  \\
  \hspace{5mm} With ASPP in KE head & 0.91 & 76.1 & 80.1 & 78.0 &  \bf \textcolor[RGB]{0,160,0}{$\downarrow$ 1.1} \\
  \hspace{5mm} With ASPP in (backbone 1/16)  & 0.89 & 73.1 & 81.3 & 77.0 &  \bf \textcolor[RGB]{0,160,0}{$\downarrow$ 2.1} \\
  \hspace{5mm} With deformable convolution (C4-1) & 0.87 & 79.5 & 83.9 & 81.7 & \bf \color{red}{$\uparrow$ 2.6} \\
  \hspace{5mm} With deformable convolution (C4-2) & 0.89 & 79.1 & 84.3 & 81.6 & \bf \color{red}{$\uparrow$ 2.5} \\
  \hspace{5mm} With deformable convolution (C3-) & 0.83 & 81.2 & 81.9 & 81.6 & \bf \color{red}{$\uparrow$ 2.5} \\
  \hspace{5mm} With panoptic segmentation (dice loss) & 0.67 & 77.7 & 80.3 & 79.0 &  \bf \textcolor[RGB]{0,160,0}{$\downarrow$ 0.1} \\
  \hspace{5mm} With pyramid attention network (PAN) & 0.85 & 77.6 & 83.1 & 80.3 & \bf \color{red}{$\uparrow$ 1.2} \\
  \hspace{5mm} With multi-scale network (MSN) & 0.91 & 79.0 & 81.6 & 80.3 & \bf \color{red}{$\uparrow$ 1.2} \\
  \noalign{\smallskip}\hline
  {\bf Proposal generation} \\
  \hspace{5mm} With deformable PSROI pooling & 0.91 & 80.7 & 79.4 & 80.0 &  \bf \color{red}{$\uparrow$ 0.9} \\
  \noalign{\smallskip}\hline
  {\bf Prediction head} \\
  \hspace{5mm} With character head & 0.93 & 77.7 & 82.0 & 79.8 &  \bf \color{red}{$\uparrow$ 0.7} \\
  \hspace{5mm} With OHEMv1 & 0.59 & 76.9 & 80.0 & 78.4 &  \bf \textcolor[RGB]{0,160,0}{$\downarrow$ 0.7} \\
  \hspace{5mm} With OHEMv2 & 0.65 & 75.8 & 81.1 & 78.3 &  \bf \textcolor[RGB]{0,160,0}{$\downarrow$ 0.8} \\
  \hspace{5mm} With OHEMv3 & 0.55 & 77.5 & 79.8 & 78.6 &  \bf \textcolor[RGB]{0,160,0}{$\downarrow$ 0.5} \\
  \hspace{5mm} With mask scoring & 0.93 & 75.7 & 81.8 & 78.6 &  \bf \textcolor[RGB]{0,160,0}{$\downarrow$ 0.5} \\
  \hspace{5mm} With cascade r-cnn (ensemble) & - & 77.7 & 80.3 & 79.0 &  \bf \textcolor[RGB]{0,160,0}{$\downarrow$ 0.1} \\
  \noalign{\smallskip}\hline
  {\bf Post-processing} \\
  \hspace{5mm} With polygonal non-maximum suppression & 0.91 & 77.2 & 82.8 & 79.9 & \bf \color{red}{$\uparrow$ 0.8} \\
  \hspace{5mm} With Key Edge RPP & 0.91 & 78.5 & 79.9 & 79.2 & \bf \color{red}{$\uparrow$ 0.1} \\
  \noalign{\smallskip}\hline
  {\bf Final model} \\
  \hspace{5mm} accumulating effective modules & 0.91 & 83.2 & 89.5 & 86.2 &  \bf \color{red}{$\uparrow$ 7.1} \\
  \noalign{\smallskip}\hline
  \end{tabular}
\end{table*}

\subsection{Ablation study of data arrangement}
Considering the image diversity and the consistency and quality of annotation, we collected a 60,000-item dataset for pretraining, which consisted of 30,000 images from the LSVT \cite{sun2019icdar} training set, 10,000 images from the MLT 2019 \cite{nayef2019icdar2019} training set, and 5,603 images from ArT \cite{chng2019icdar2019}, which contained all the images from SCUT-CTW1500 \cite{liu2019curved} and Total-text \cite{kheng2017total,ch2019total}. The remaining 14,859 images were selected from RCTW-17 \cite{shi2017icdar2017}, ICDAR 2015 \cite{karatzas2015icdar}, ICDAR 2013 \cite{Karatzas2013ICDAR}, MSRA-TD500 \cite{Yao2012Detecting}, COCO-Text \cite{veit2016coco}, and USTB-SV1K \cite{Yin2015Multi}. Note that we transferred polygonal annotations to the minimum area rectangle for training.

The ablation results are presented in Table \ref{tab:Ablation}. If we only were to use the pretrained data without the split training data from the ReCTS, the result in the ReCTS validation set would be significantly worse than that of the baseline, even if the pretrained model were trained with more iterations. This is because the diversity and annotation granularity of the selected pretrained dataset is still very different from that of the ReCTS dataset. However, using the model trained with pretrained data is better than using the ImageNet model. For example, when directly using the ImageNet ResNet-101 model instead of the MLT pretrained model from the baseline method, the Hmean is reduced by 0.5\%. Using the model having 60,000 pretrained data, followed by finetuning the model on the split ReCTS training data improved the result by 1.2\% in terms of Hmean. To evaluate the importance of the data quality, we mimicked the manual annotation error by removing 5\% of the training annotation instances and did not correct some samples with annotation ambiguity from the original ReCTS training data. The results indicate that using defective training data significantly degrades the performance.

\subsection{Ablation study of pre-processing}
\label{subsec:ab_pre}
Our baseline model used a pretrained model having only a flip strategy for data augmentation. We compared the baseline with various other data augmentation methods.

\begin{figure}[t!]
	\centering
	\subfigure[Ablation results of rotated pseudo samples.]{\includegraphics[width = 8.6cm, height = 4.5cm]{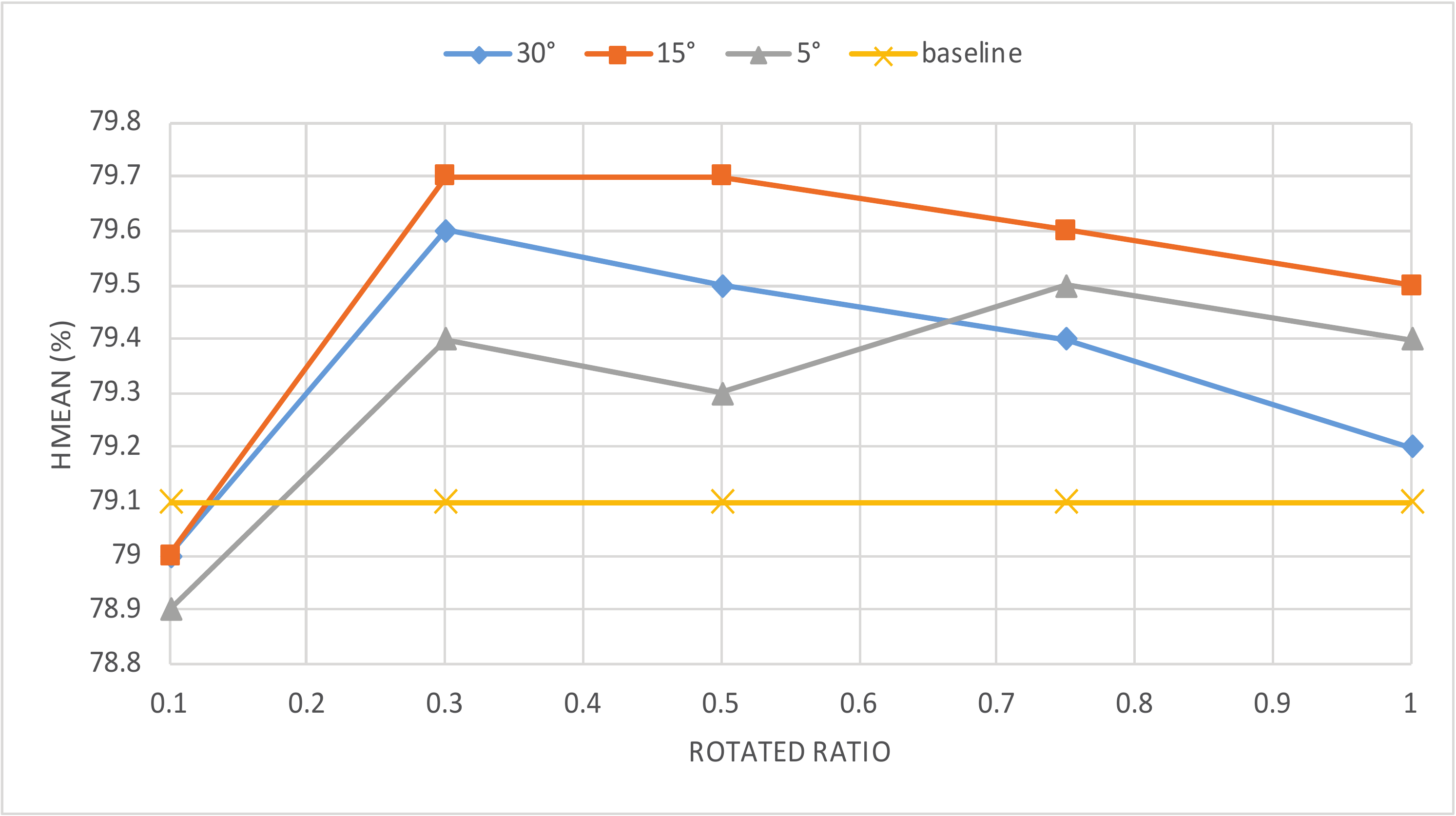}
	\label{fig:rores}}
	\subfigure[Ablation  results of crop pseudo samples.]{\includegraphics[width = 8.6cm, height = 4.5cm]{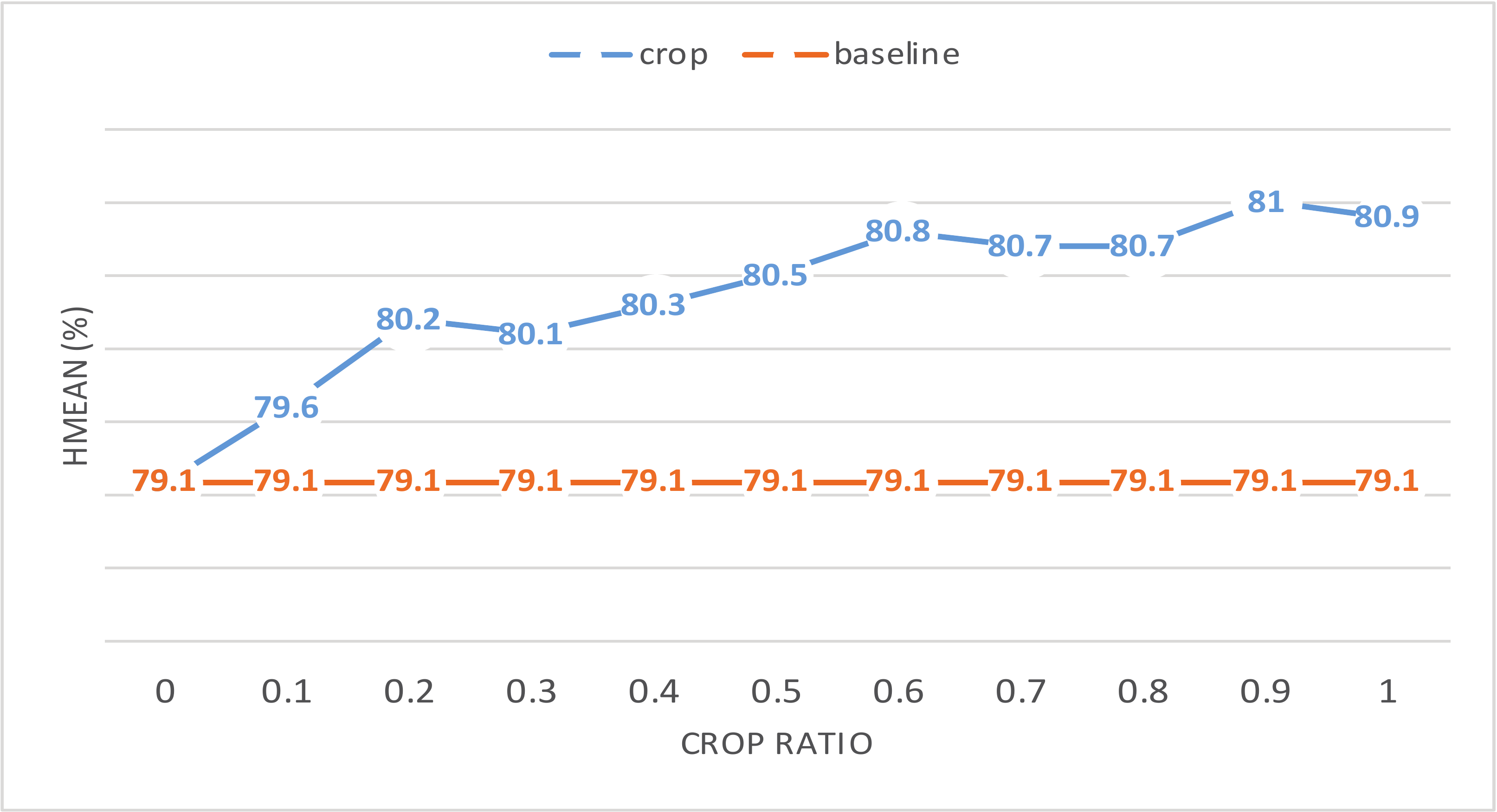}
	\label{fig:crres}}
	\caption{Ablation studies of data augmentation strategies.}
	\label{fig:rc_results}
\end{figure}

\paragraph{Cropping and rotation.} Without introducing extra parameters or training/testing times, the results presented in Table \ref{tab:Ablation} verify that both rotation and data cropping augmentation strategies improved the detection results. We further conducted a sensitivity analysis of how the ratios of using these two strategies influence the performance, as shown in Figure \ref{fig:rc_results}. Some useful findings can be derived from Figure \ref{fig:rores}, as summarized below.
\begin{itemize}
  \item With appropriate ratios, three rotated degrees (30$^{\circ}$, 15$^{\circ}$, and 5$^{\circ}$) outperformed the baseline method in most ratios, with 0.5, 0.6, and 0.4\%, respectively. 
  \item Under a 0.1 rotated ratio, the performances with the three rotated degrees were all worse than the baseline. This may be because the pseudo samples changed the distribution of the original dataset, whereas very few pseudo samples were insufficient to improve the generalization ability. Conversely, the ratios to achieve the best results for various rotated degrees always lie between 0.3 and 0.8, which empirically suggests that using a medium ratio for the rotated data augmentation strategy might be a suitable choice.
  \item We can also see that the performance using a rotated angle of 15$^{\circ}$ was consistently better than that with 30$^{\circ}$ and 5$^{\circ}$.
\end{itemize}

Compared with the rotated data augmentation strategy, the random cropping strategy significantly improved detection performance. The best performance, as shown in Table \ref{tab:Ablation}, achieved a 1.9\% improvement in terms of Hmean, compared with the baseline method. Sensitivity analysis, as shown in Figure \ref{fig:crres}, was also conducted, revealing that, as the crop ratio improved, the performance also tended to improve. The result suggests that always using the crop strategy was conducive to improving the detection results. Note that a crop ratio of 0.1 only improved the Hmean by 0.5\%, whereas other ratios improved it by more than 1\%, which is similar to the phenomenon when using a rotated ratio 0.1.

\paragraph{Color jittering.} We also conduct a simple ablation study to evaluate the performance of color jittering. Based on the same settings as of the baseline method, we empirically set the ratios of brightness, contrast, saturation, and hue to 0.5, 0.5, 0.5, and 0.1, respectively. The ratio represents the degree of disturbance of each specific transformation. The results in Table \ref{tab:Ablation} indicate that using color jittering data augmentation slightly improved the result by 0.2\% in terms of Hmean.

\paragraph{Training image scale.} The training image scale/size is specifically important for a scene text detection. To evaluate how the training scale influences the results of our method, we used two parameters (\textit{i.e.}, $scale$ and $MaxSize$) to control the training scale. The first item resized the minimum side of the image to a specific parameter. 

In our implementation, there are a set of values for random scaling. The second item restricts the maximum size of the image sides. The value of $scale$ must be less than $MaxSize$, and the entire scaling process strictly retains the original aspect ratio. We primarily compare three different settings: the default training scale ($scale$: 560 to 920 with intervals of 40, $MaxSize$ was 1,300); medium training scale ($scale$: 680 to 1,120 with intervals of 40, $MaxSize$ was 1,800); and large training scale ($scale$ 800 to 1,400 with intervals of 40, $MaxSize$ was 2,560). 

The results are presented in Table \ref{tab:big_scale}, which verify the following: 1) a larger training scale requires a larger testing scale for the best performance. 2) As the larger training scale increases, so does the performance. Note that, although a larger training scale can improve performance, it is costly and may require significantly more GPU memory.

\begin{table}
  \caption{Ablation experiments for large scale training. Hmean$_{1}$, Hmean$_{2}$, and Hmean$_{3}$ represent default training scale, medium training scale, and large training scale, respectively. The first row compares the performance based on the baseline setting. The other three rows are the best setting (using grid search to find the best $scale$ and $MaxSize$) for each training scale. }
  \label{tab:big_scale}       %
  \resizebox{0.47\textwidth}{!}{
  \begin{tabular}{cccc}
  \hline\noalign{\smallskip}
  ($Scale$, $MaxSize$) & Hmean$_{1}$ (\%) & Hmean$_{2}$ (\%) & Hmean$_{3}$ \\
  \noalign{\smallskip}\hline\noalign{\smallskip}
  (1400, 2000) & 79.1 & 81.3 & 81.9 \\
  \hline
  (800, 1300) & 81.5 & - & - \\
  (1600, 1600) & - & 82.2 & - \\
  (1600, 1700) & - & - & 82.5 \\
  \noalign{\smallskip}\hline
  \end{tabular}
  }
\end{table}

\subsection{Ablation study of the backbone}
A well-known hypothesis is that a deeper and wider network architecture delivers better performance than does a shallower and thinner one. However, increasing the network depth naively will significantly increase the computational cost with only limited improvement. Therefore, we investigate different %
types
of backbone architectures. The results are shown in Table \ref{tab:Ablation} and are summarized as follows:
\begin{itemize}
  \item By changing the backbone, ResNet-101-FPN of the baseline model into a ResNeXt-152-32x8d-FPN-IN5k, Hmean can be increased by 2.5\%. Note that the pretrained model of ResNeXt-152-32x8d-FPN-IN5k was pretrained on ImageNet using the Facebook Detectron framework. 
  \item Atrous spatial pyramid pooling (ASPP) \cite{chen2017deeplab} is effective in semantic segmentation, which is known for its function in increasing the receptive field. However, in this scene text detection task, using ASPP in the KE head or backbone reduced  performance by 1.1 and 2.1\%, respectively. One possible reason is that the change in network architecture usually requires more iterations. However, the best confidence thresholds for the best performance using ASPP were 0.91 and 0.89, which are similar to the best threshold of the baseline model, suggesting that the network had already converged.
  \item Deformable convolution \cite{dai2017deformable} is an effective module used for many tasks. It adds 2D offsets to the regular sampling grid of the standard convolution, allowing free form deformation of the convolutional operation. This is suitable for scene text detection, owing to the mutable characteristics of the text. We experimented with three methods of deformable convolution by adding deformable convolutions from the C4-1, C4-2, and C3 of the backbone, and the results show that the performance could be significantly improved by 2.6, 2.5, and 2.5\%, respectively, in terms of Hmean.
  \item Motivated by the panoptic feature pyramid networks \cite{kirillov2019panoptic}, we also tested whether a panoptic segmentation loss was useful for scene text detection. To this end, we used a dice loss in the output of the FPN for panoptic segmentation, which had two classes: background and text. The result in Table \ref{tab:Ablation} indicates that Hmean was reduced by 0.1\%. However, the best threshold was 0.67, which indicates that the background noise may have somehow reduced the confidence of the training procedure.
  \item The pyramid attention network (PAN) \cite{huang2019mask} is a novel structure that combines an attention mechanism and a spatial pyramid to extract precise dense features for semantic segmentation tasks. Because it can effectively suppress false alarms caused by text-like backgrounds, we integrated it into the backbone and tested its function. The results show that using PAN led to a 1.2\% improvement in terms of Hmean, but it also increased the computational cost with an increase of 2.4 GB video memory.
  \item The multi-scale network (MSN) \cite{xue2019msr} is robust for scene text detection because it employs multiple network channels to extract and fuse features at different scales concurrently. In our experiment, integrating MSN into the backbone also increased the performance by 1.2\% in terms of Hmean. Note that, compared with PAN, the recall of the MSN was much better under a higher best threshold, which suggests that different architectures may have had different functions related to the performance of the detector.
\end{itemize}

\subsection{Ablation study on proposal generation}
The proposed model is based on a two-stage framework, and the region proposal network (RPN) \cite{ren2015faster} is used as the default proposal generation mechanism. 

Previous studies have modified the anchor generation mechanism, including DMPNet \cite{liu2017deep}, DeRPN \cite{xie2018derpn}, Kmeans anchor \cite{redmon2017yolo9000}, scale-adaptive anchor \cite{li2019scale}, and guided anchor \cite{wang2019region}, to improve the results. For simplicity, we retrain  the default RPN structure with the statistical setting of the anchor box based on the training set. 

The other important part in this proposal generation stage is the sampling process, (e.g., RoI pooling \cite{ren2015faster}, RoI align \cite{he2017mask} (our default setting), and PSRoI pooling \cite{dai2016r}. We choose to evaluate Deformable PSRoI Pooling \cite{dai2017deformable} for our method, because it has been effective for scene text detection \cite{yang2018inceptext}, and the flexible process may be beneficial to the proposed \Ours. The result is shown in Table \ref{tab:Ablation}: using deformable PSRoI Pooling improved the baseline method by 0.9\% in terms of Hmean.

\begin{table}
  \caption{Ablation results of using cascade r-cnn. cf: best threshold. R: recall. P: precision. H: Hmean.}
  \label{tab:cascade}       %
  \resizebox{0.47\textwidth}{!}{
  \begin{tabular}{cccccc}
  \hline\noalign{\smallskip}
  Method & cf & R (\%) & P (\%) & H (\%) & $\Delta$H \\
  \noalign{\smallskip}\hline\noalign{\smallskip}
  Baseline model & 0.91 & 78.1 & 80.1 & 79.1 & - \\
  \hline
  Stage 1 & 0.91 & 74.7 & 81.8 & 78.1 & \bf \textcolor[RGB]{0,160,0}{$\downarrow$ 1.0} \\
  Stage 2 & 0.87 & 76.3 & 81.1 & 78.6 & \bf \textcolor[RGB]{0,160,0}{$\downarrow$ 0.5} \\
  Stage 3 & 0.87 & 75.9 & 79.5 & 77.7 & \bf \textcolor[RGB]{0,160,0}{$\downarrow$ 1.4} \\
  ensemble & - & 77.7 & 80.3 & 79.0 & \bf \textcolor[RGB]{0,160,0}{$\downarrow$ 0.1} \\
  \noalign{\smallskip}\hline
  \end{tabular}
  }
\end{table}

\subsection{Ablation study on the prediction head}
The final part of the two-stage detection framework is the prediction head. To clearly evaluate the effectiveness of the components, ablation experiments are separately conducted on different heads.

\paragraph{Box head.} 
Empirically, online hard negative examples mining (OHEM) \cite{shrivastava2016training} is not always effective with respect to different benchmarks. For example, using the same framework minus the training data can significantly improve the results with the ICDAR 2015 benchmark \cite{karatzas2015icdar} while reducing the results on the MLT benchmark \cite{nayef2017icdar2017}. This finding may be related to the data distribution, which is difficult to trace. 

Thus, we test three versions of the OHEM in the validation set. The first version, OHEMv1, is the same as the original implementation; the second version, OHEMv2, simply ignores the top 5 hard examples to avoid outliers.
These two versions have the same ratio, which is set to 0.25. The third version, OHEMv3, simply uses a higher ratio (0.5) to guarantee more hard samples and less easy samples. The results in Table \ref{tab:Ablation} show that three versions all reduce Hmean, by 0.7, 0.8, and 0.5, respectively. Note that using OHEM will also result in the reduction of the best confidence, which means that the forced learning of hard examples can reduce the confidence of normal examples. Conversely, we also evaluated the performance of the cascade R-CNN, and the results are shown in Table \ref{tab:cascade}. However, the results show that using a cascade does not result in further improvements.

\paragraph{Mask head.} To improve the mask head, we evaluate two methods (\textit{i.e.}, mask scoring \cite{huang2019mask}), as shown in Table \ref{tab:Ablation}. The results show that modification of the mask head does not contribute to the detection performance. However, the mask prediction results are visually more compact and accurate compared with the baseline. 
\begin{table}[t!]
  \caption{Ablation  experiments for using character head. H: Hmean.}
  \label{tab:character}       %
  \centering
  \begin{tabular}{ccc}
  \hline\noalign{\smallskip}
  Method & H (\%)  & $\Delta$H\\
  \noalign{\smallskip}\hline\noalign{\smallskip}
  Baseline & 79.1 & - \\
  \hline
  Baseline + character head  & 79.8 & \bf \color{red}{$\uparrow$ 0.7} \\
  \makecell*[c]{Baseline + character head \\+ mask character} & 79.8 & \bf \color{red}{$\uparrow$ 0.7} \\
  \makecell*[c]{Baseline + character head \\+ instance connection} & 79.6 & \bf \color{red}{$\uparrow$ 0.5} \\
  \makecell*[c]{Baseline + character head \\+ instance connection \\ - KE head} & 75.2 & \bf \textcolor[RGB]{0,160,0}{$\downarrow$ 3.9} \\
  \noalign{\smallskip}\hline
  \end{tabular}
\end{table}

\begin{figure*}[!t]
  \centering
  \centerline{\includegraphics[width=17.6cm, height = 6cm]{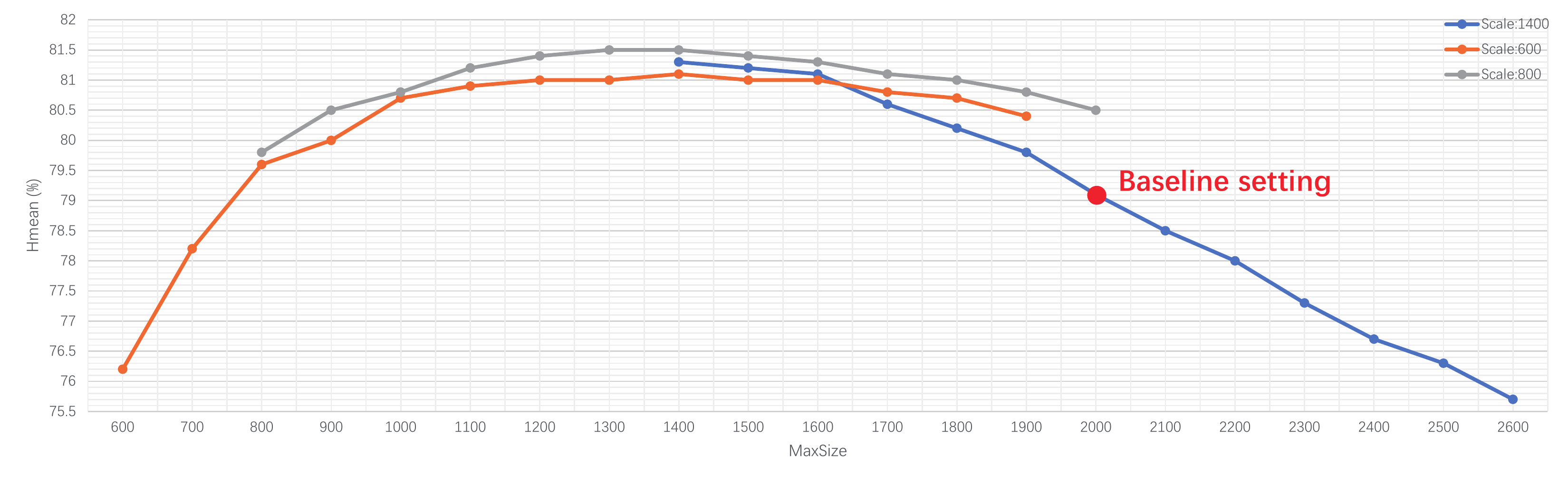}}
  \caption{Ablation study of the testing scale. Note that the training scale is the default setting mentioned in Section \ref{subsec:ab_pre}.}\label{fig:test_scale}
\end{figure*}

\paragraph{Character head.} 
It is well known
that stronger supervision can result in better performance. Because the competition also provides a character ground truth, we build and evaluate  the performance of an auxiliary character head. The implementation of the character head is exactly the same as that for the box head, except for the ground truth. Unlike the box, mask, and KE head, the proposed character head is built on a different RPN. Thus, the character head does not share the same proposal with the other heads. The KE head directly produces 
a quadrilateral bounding box (word box) directly used for the final detection, and we test whether the auxiliary head could indirectly (shared backbone) improve the word-box detection performance. 

The ablation results in Table \ref{tab:character} demonstrate this idea, which shows that using a character head improved the Hmean by 0.7\%. Additionally, if we %
add a mask prediction head to the character head (\textit{i.e.}, the mask character in Table \ref{tab:character}), the result would remain the same. Moreover, we employ a triplet loss to learn the connection between the characters. The ground truth includes whether the characters belong to the same text instances. However, the improvement is decreased to 0.5\%. This may be because the instance connection introduced an inconsistent labeling issue. We further test the performance using only the character head with an instance connection and without the KE head. Hmean is reduced by 3.9\% compared with the baseline method, suggesting that using character as an auxiliary head instead of the final prediction head %
is a good choice.

\subsection{Ablation study of post-processing}
\label{subsec:ab_post}
The last step is to apply post-processing methods for final improvement. To this end, we compare the baseline with a series of standard and more effective post-processing methods.

\paragraph{Polygonal non-maximize suppression (PNMS).} Traditional non-maximum suppression (NMS) methods between horizontal rectangular bounding boxes can cause unnecessary suppression. Thus, we conduct ablation experiments to evaluate the performance of the PNMS. We use  grid search to find the best threshold to find both NMS and PNMS for fair combination, which is 0.3 and 0.15, respectively. The result in Table \ref{tab:Ablation} shows that using PNMS performs better than NMS by 0.8\% in terms of Hmean. Additionally, PNMS %
is
much more effective when using a test ensemble in practice.

\paragraph{Key edge RPP.} The proposed key edge RPP proved effective on the ICDAR 2015 benchmark. Thus, we also test whether it %
applies 
to the competition dataset. The ablation result in Table \ref{tab:Ablation} shows that it slightly improves  the Hmean by 0.1\% compared with the baseline. It is worth noticing that, although the best confidence threshold  is 0.91, which is the same as that of the baseline, the recall is increased by 0.4\% while only reducing the precision by 0.2\%.

\paragraph{Large-scale testing.} We also conduct  experiments to evaluate how the testing scale influenced performance. The results are shown in Figure \ref{fig:test_scale}, which demonstrates that a proper setting of $scale$ and $MaxSize$ significantly improves the detection performance. Additionally, the results reveal that there is a limitation of the $MaxSize$. That is, if the value of $MaxSize$ is higher than a certain value, the performance would be gradually reduced.

\paragraph{Test ensemble.} To evaluate the performance of the test ensemble, we conduct ablation experiments with four different aspects: different backbone ensemble; multiple intermediate model ensemble; a multi-scale ensemble; and an independent model ensemble. Note that, to achieve the best performance, implementing ensemble or multi-scale testing requires some tricks. Otherwise, the results may be worse. We summarize the results as follows: 
\begin{itemize}
  \item Using a high confidence threshold. One weakness of multi-scale ensembling is that if a true-negative detection exists in one of the testing scales, it cannot be avoided unless we set a high confidence threshold to exclude it during the ensemble phase. Therefore, for each scale, we first test its best confidence threshold (cf) on the validation set. Then, we use a higher confidence for the model ensemble.
  \item Variant scale of multi-scale testing. The performance of small scale (600 ($scale$), 1200 ($MaxSize$)) is rated. For example, in the ReCTs competition, it is much worse than that of large-scale (1,600, 1,600). However, small scales are %
  better for
  detecting large instances compared with large scales, and they can always be mutually promoted in practice.
  \item Using a strict PNMS threshold. A normal case for the ensemble result is that the recall can be significantly improved, whereas the prediction is dramatically reduced. When observing the final integrated detection boxes, it is easy to find that the reduction was caused by boxes-in-boxes and many stacked redundant boxes. Using a strict PNMS can effectively solve this issue.
\end{itemize}

\begin{table*}[!t]
  \caption{Ablation  experiments for different approaches of model ensemble. `def': deformable convolution.}
  \label{tab:model_ensemble}
  \centering
  \newcommand{\tabincell}[2]{\begin{tabular}{@{}#1@{}}#2\end{tabular}}
  \small
  \resizebox{\textwidth}{!}{
  \begin{tabular}{c||c|c|c||c|c|c||c|c|c||c|c}
    \hline
    Method & \multicolumn{3}{c||}{Backbone ensemble} & \multicolumn{3}{c||}{Intermediate model ensemble} & \multicolumn{3}{c||}{Multi-scale ensemble ($scale$, $MaxSize$)} & \multicolumn{2}{c}{Model ensemble} \\
    \hline
    Components &  def-C4-1 & def-C4-2 & def-C3 & x152-60k & x152-70k & x152-80k  & (600,1600) & (1200,1600) & (1600,1600) & M1 & M2  \\
    \hline
    Hmean (\%)  & 81.7 & 81.6  & 81.6 & 80.7 & 81.7  & 81.6 & 79.8 & 82.1 & 82.6 & 83.2 & 83.5 \\
    \hline
    Ensemble &  \makecell*[c]{def-C4-1 \\ \& def-C4-2} & \makecell*[c]{def-C4-1 \\ \& defC3} & \makecell*[c]{def-C4-1 \\ \& def-C4-2 \\ \& def-C3}  & \multicolumn{3}{c||}{x152-60k \& x152-70k \& x152-80k} & \multicolumn{3}{c||}{(600, 1600) \& (1200, 1600) \& (1600, 1600)} & \multicolumn{2}{c}{M1 \& M2}\\
    \hline
    Hmean (\%)  & 81.8 & 82.1 & 82.2 & \multicolumn{3}{c||}{81.9} & \multicolumn{3}{c||}{83.2} & \multicolumn{2}{c}{83.7}\\
    \hline
  \end{tabular}
  }
  \end{table*}

\begin{table}[!t]
    \caption{Experimental results for the ICDAR 2015 dataset. R: recall. P: precision. }
    \label{tab:ic15}
    \centering
    \newcommand{\tabincell}[2]{\begin{tabular}{@{}#1@{}}#2\end{tabular}}
    \small
    \begin{tabular}{ r  |ccc}
      \hline
      Algorithms  & $R (\%)$  & $P (\%)$ & $Hmean (\%)$ \\
      \hline
      \hline
      Tian et al. \cite{tian2016detecting} & 52.0 & 74.0 & 61.0 \\
      \hline
      Shi et al. \cite{shi2017detecting} & 76.8  & 73.1 & 75.0\\
      \hline
      Liu et al. \cite{liu2017deep} & 68.2 & 73.2 & 70.6\\
      \hline
      Zhou et al. \cite{zhou2017east}  & 73.5 & 83.6 & 78.2\\
      \hline
      Ma et al. \cite{ma2018arbitrary}  & 73.2 & 82.2 & 77.4\\
      \hline
      Hu et al. \cite{hu2017wordsup} & 77.0 & 79.3 & 78.2\\
      \hline
      Liao et al. \cite{liao2018rotation} & 79.0 & 85.6 & 82.2\\
      \hline
      Deng et al. \cite{deng2018pixellink} & 82.0 & 85.5 & 83.7\\
      \hline
      Ma et al. \cite{ma2018arbitrary} & 82.2 & 73.2 & 77.4\\
      \hline
      Lyu et al. \cite{lyu2018multi} & 79.7 & 89.5 & 84.3\\
      \hline
      He et al. \cite{he2017deep} & 80.0 & 82.0 & 81.0 \\
      \hline
      Xu et al. \cite{xu2019textfield} & 80.5 & 84.3 & 82.4  \\
      \hline
      Tang et al. \cite{tang2019detecting} & 80.3 & 83.7 & 82.0  \\
      \hline
      Wang et al. \cite{wang2019shape} & 84.5 & 86.9 & 85.7 \\
      \hline
      Xie et al. \cite{xie2018scene} & 85.8 & 88.7 & 87.2 \\
      \hline
      Zhang et al. \cite{zhang2019look} & 83.5 & 91.3 & 87.2  \\
      \hline
      Liu et al. \cite{liu2018fots} & 87.9 & 91.9 & 89.8 \\
      \hline
      Baek et al. \cite{baek2019character} & 84.3 & 89.8 & 86.9 \\
      \hline
      Huang et al. \cite{huang2019mask1} & 81.5 & 90.8 & 85.9 \\
      \hline
      Zhong et al. \cite{zhong2019improved} & 80.1 & 87.8 & 83.8 \\
      \hline
      He et al. \cite{he2018end2} & 86.0 & 87.0 & 87.0 \\
      \hline
      Liu et al. \cite{liu2019arbitrarily} & 87.6 & 86.6 & 87.1 \\
      \hline
      Liao et al. \cite{liao2018textboxes++} & 78.5 & 87.8 & 82.9 \\
      \hline
      Long et al. \cite{long2018textsnake} & 80.4 & 84.9 & 82.6 \\
      \hline
      He et al. \cite{he2020realtime} & 79.7 & 92.0 & 85.4 \\
      \hline
      Lyu et al. \cite{lyu2018mask} & 81.0 & 91.6 & 86.0 \\
      \hline
      He et al. \cite{he2017single} & 73.0 & 80.0 & 77.0 \\
      \hline
      Wang et al. \cite{xie2019convolutional} & 79.6 & 83.2 & 81.4 \\
      \hline
      Liao et al. \cite{liao2019mask} & 87.3 & 86.6 & 87.0 \\
      \hline
      Wang et al. \cite{wang2019efficient} & 81.9 & 84.0 & 82.9 \\
      \hline
      Wang et al. \cite{Wang_2019_CVPR} & 86.0 & 89.2 & 87.6 \\
      \hline
      Qin et al. \cite{qin2019towards} & 88.0 & 91.7 & 89.8 \\
      \hline
      Feng et al. \cite{feng2019textdragon} & 83.8 & 92.5 & 87.9 \\
      \hline\hline
      Liu et al.\  \cite{liu2019omnidirectional} & 83.8 & 89.4 & 86.5 \\
      \hline
      Ours & 88.2 & 92.1 & \bf 90.1 \\
      \hline
    \end{tabular}
   \end{table}

Based on these principles, we conclude the results of the four ensemble aspects as follows.
\begin{itemize}
  \item {\bf Different backbone ensembles.} We train three models using the baseline setting with three types of deformable convolution, starting from C4-1, C4-2, and C3 of ResNet-101, respectively. The ensemble results of the three methods are shown in Table \ref{tab:model_ensemble}. From the table, we can see that integrating the models with a series of simple backbone modifications improved the detection performance, even based on a relatively high baseline. Additionally, the results show that integrating more components resulted in better performance.
  \item {\bf Multiple intermediate model ensembles.} We also evaluate the performance of integrating intermediate models. We use the trained model with the ResNext-152 backbone as a strong baseline and selected the last three intermediate iterating models with 10,000 iterations as intervals for the ensemble. The results shown in Table \ref{tab:model_ensemble} also demonstrate that when using the model ensemble, the intermediate models could be mutually promoted.
  \item {\bf Multi-scale ensemble.} To evaluate the performance of the multi-scale ensemble, we use grid searching to find the best PNMS threshold for three specified settings ($scale$, $MaxSize$), representing large, medium, and small text instances, respectively. Each detection result was then integrated with a PNMS threshold 0.02 higher than the original best threshold, which resulted in approximate optimum integrating results with 0.6\% improvement in terms of Hmean, as shown in Table \ref{tab:model_ensemble}. 
  \item {\bf Independent model ensembles.} Finally, we test the performance of integrating the two final models. The first model contains the baseline setting plus deformable convolution, and the second model contains the baseline setting with the ResNext-152 backbone. We independently integrate each model using an intermediate model ensemble and a multi-scale ensemble. Then, we assemble the final results of the two models. As shown in Table \ref{tab:model_ensemble}, the detection result can still be improved.
\end{itemize}

   \begin{table}[!t]
    \caption{Experimental results for the MLT dataset. SS represents a single scale. R: recall. P: precision. Note that we only used a single scale for all experiments. }
    \label{tab:mlt}
    \centering
    \newcommand{\tabincell}[2]{\begin{tabular}{@{}#1@{}}#2\end{tabular}}
    \small
    \begin{tabular}{r |ccc}
      \hline
      Algorithms  & $R (\%)$  & $P (\%)$ & $Hmean (\%)$ \\
      \hline
      \hline
      linkage-ER-Flow  \cite{nayef2017icdar2017} & 25.59 & 44.48 & 32.49 \\
      \hline
      TH-DL  \cite{nayef2017icdar2017} & 34.78 & 67.75 & 45.97 \\
      \hline
      SARI FDU RRPN v2 \cite{ma2018arbitrary} & 67.0 & 55.0 & 61.0 \\
      \hline
      SARI FDU RRPN v1 \cite{ma2018arbitrary} & 55.5 & 71.17 & 62.37 \\
      \hline
      Sensetime OCR \cite{nayef2017icdar2017} & 69.0 & 67.75 & 45.97 \\
      \hline
      SCUT\_DLVClab1 \cite{liu2017deep} & 62.3 & 80.28 & 64.96 \\
      \hline
      AF-RNN  \cite{zhong2018anchor} & 66.0 & 75.0 & 70.0 \\
      \hline
      Lyu et al. \cite{lyu2018multi} & 70.6 & 74.3 & 72.4 \\
      \hline
      FOTS \cite{liu2018fots} & 62.3 & 81.86 & 70.75 \\
      \hline
      CRAFT \cite{baek2019character} & 68.2 & 80.6 & 73.9 \\
      \hline\hline
      Liu et al.\  \cite{liu2019omnidirectional}  & 70.1 & 83.6 & 76.3 \\
      \hline
      Ours & 76.44 & 82.75 & \textbf{79.47} \\
      \hline
    \end{tabular}
    \end{table}
  
    \begin{table*}
      \centering
      \small 
      \caption{Competition results on the ReCTS dataset. The results are from the competition website
      {\color{blue}
      \url{https://tinyurl.com/ReCTS2019}. 
      } For the detection task, the ranking is based on Hmean. For End-to-End detection and recognition task, the ranking is based on 1-NED. NED: normalized edit distance.}
      \label{tab:rects}       %
        \begin{tabular}{ r |ccc|cccc}
        \hline\noalign{\smallskip}
        \multirow{2}*{Affiliation} & \multicolumn{3}{c|}{Detection Result} & \multicolumn{4}{c}{End-to-End Result} \\
        \cline{2-8}
              & Recall (\%) & Precision (\%) & Hmean (\%) & Recall (\%) & Precision (\%)& Hmean (\%) & 1-NED (\%) \\
        \hline
        \bf Ours  & 93.97  & 92.76 & \bf 93.36 & 93.97  & 92.76 & 93.36  & \bf 81.62 \\
        \hline
        Tian et al. & 93.46  & 92.59& 93.03& 92.49 & 93.49 & 92.99  & 81.45 \\
        Liu et al.& 93.41  & 91.62  & 92.50& -  & - & -  & - \\
        Zhu et al.& 93.51  & 89.15 & 91.27& 92.36  & 91.87 &  92.12  & 79.38 \\
        Mei et al.& 91.96  & 90.09 & 91.02& -  & - & -  & - \\
        Li et al.& 90.03  & 91.65 & 90.83& 90.80  & 90.26 & 90.53  & 73.43 \\
        Zheng et al.& 89.84  & 91.41 & 90.62& -  & - & -  & - \\
        Zhou et al.& 90.99  & 89.59 & 90.28& 90.99  & 89.59 & 90.28 & 74.35 \\
        Zhang et al.& 93.66  & 86.35 & 89.86& 93.62  & 87.22 & 90.30  & 76.60 \\
        Zhao et al. & 86.13  & 92.72 & 89.31& 86.12  & 92.73 & 89.30  & 72.76 \\
        Xu et al. & -  & - & - & 91.54  & 90.28 & 90.91  & 71.89 \\
        Wang et al. & 88.92  & 88.70 & 88.80& 88.89  & 88.92 & 88.91  & 71.81 \\
        Baek et al. & 85.33  & 89.38 & 87.31& 75.89  & 78.44 & 77.14  & 41.68 \\
        Wang et al. & 84.67  & 89.53 & 87.03& 84.64  & 89.56 & 87.03  & 71.10 \\
        Wang et al. & -  & - & - & 69.49  & 89.52 & 78.24  & 50.36 \\
        Li et al.& 82.27  & 88.49 & 85.27& -  & - & -  & - \\
        Xu et al.& 88.52  & 79.32 & 83.66& -  & - & -  & - \\
        Lu et al.& 85.18  & 79.66 & 82.33& -  & - & -  & - \\
        Ma et al.& 83.16  & 80.77 & 81.94& -  & - & -  & - \\
        Tian et al.& 96.17  & 69.20 & 80.48& -  & - & -  & - \\
        Feng et al.& 73.05  & 78.35 & 75.61& -  & - & -  & - \\
        Luan et al.& 70.35  & 80.19 & 74.95& -  & - & -  & - \\
        Yang et al.& 60.66  & 90.87 & 72.76& -  & - & -  & - \\
        Liu et al.& 66.83  & 75.87  & 71.07& -  & - & -  & - \\
        Zhou et al.& 72.54  & 56.44 & 63.48& -  & -& -  & - \\
        Liu et al.& 7.82  & 8.14 & 7.98& -  & - & -  & - \\
        \noalign{\smallskip}\hline
        \end{tabular}
    \end{table*} 
  
  \section{Comparison with state-of-the-art methods}
  To further evaluate the effectiveness of the proposed method, we carry out experiments and compare our final model with other state-of-the-art methods on three scene text datasets: ICDAR 2015 \cite{karatzas2015icdar}, MLT \cite{nayef2017icdar2017}, and ReCTS (See Section \ref{subsubsec:rects_cd}). We also conduct an experiment on one aerial dataset, HRSC2016 \cite{liu2017rotated}, to further demonstrate the generalization ability of our method.
  
  {\bf Final model.} The final model is designed by %
  combine
  the effective modules evaluated in Table \ref{tab:Ablation}. Specifically, based on the baseline setting, we refine our model in all six aspects. During the data arrangement stage, we use 60,000 pretrained data items to train a pretrained model for 200,000 iterations, and we then use the original training data of each dataset for finetuning. In the pre-processing part, apart from the baseline setting, we also apply color jittering, random cropping, and random rotation with their best ratios as evaluated on the validation dataset for data augmentation. Additionally, the images are trained with a medium setting of the random scale training for maximizing the utilization of the video memory. For the backbone setting, we integrate the ResNext-152-32x8d-FPN-IN5k model, deformable convolution (C4-2), PAN, and MSN modules together to construct a powerful feature extractor. During the proposal generation stage, we adopt deformable PSROI pooling for feature alignment, whereas in the prediction head, we only add an auxiliary character head for mutual promotion using only the ReCTS dataset. Finally, in the post-processing stage, we utilize all effective settings, including polygonal non-maximum suppression, key edge RPP, intermediate model ensemble, and multi-scale ensemble.

  {\bf The ICDAR 2015 Incidental Scene Text} \cite{karatzas2015icdar} is one of the most popular benchmarks for oriented scene text detection. The images are incidentally captured mostly from streets and shopping malls. Thus, the challenges of this dataset rely on oriented, small, and low-resolution text. This dataset contains 1,000 training samples and 500 testing samples with approximately 2,000 content-recognizable quadrilateral word-level bounding boxes. The results of ICDAR 2015 are given in Table \ref{tab:ic15}. From this table, it is clear that our method outperformed all previous methods.

  {\bf The ICDAR 2017 MLT} \cite{nayef2017icdar2017} is the largest multi-lingual (nine languages) oriented scene text dataset, including 7,200 training samples, 1,800 validation samples, and 9,000 testing samples. The challenges associated with this dataset are manifold. Different languages have different annotating styles. For example, most Chinese annotations are long, and there is no specific word interval for sentences. However, most English annotations are short. The annotations of Bangla or Arabic may be frequently entwined with each other, and there is more multi-orientation, perspective distorted text on various complex backgrounds. Furthermore, many images have more than 50 text instances. All instances are well annotated with compact quadrangles. As shown in Table~\ref{tab:mlt}, the proposed approach achieved the best performance on the MLT dataset.
  
  {\bf ReCTS} is the recent ICDAR 2019 Robust Reading Challenge\footnote{\url{https://rrc.cvc.uab.es/?ch=12&com=introduction}} described in Section \ref{subsubsec:rects_cd}. Competitors were restricted to submitting at most five results, and all results were evaluated after the deadline. The competition attracted numerous competitors from well-known universities and high-tech companies. The results of the ReCTS are shown in Table \ref{tab:rects}. Our method won first place in the ReCTS detection competition. 
  To clearly evaluate the performance of the final model, we also provide the results of our method on the ReCTS validation set without using a model ensemble. As shown in Table \ref{tab:Ablation}, the final model significantly outperformed the baseline by 7.1\% in terms of Hmean.
  
\begin{figure*}[t!]
    \centering
    \subfigure[Detection only results on MLT dataset.]{\includegraphics[width=0.95\textwidth]{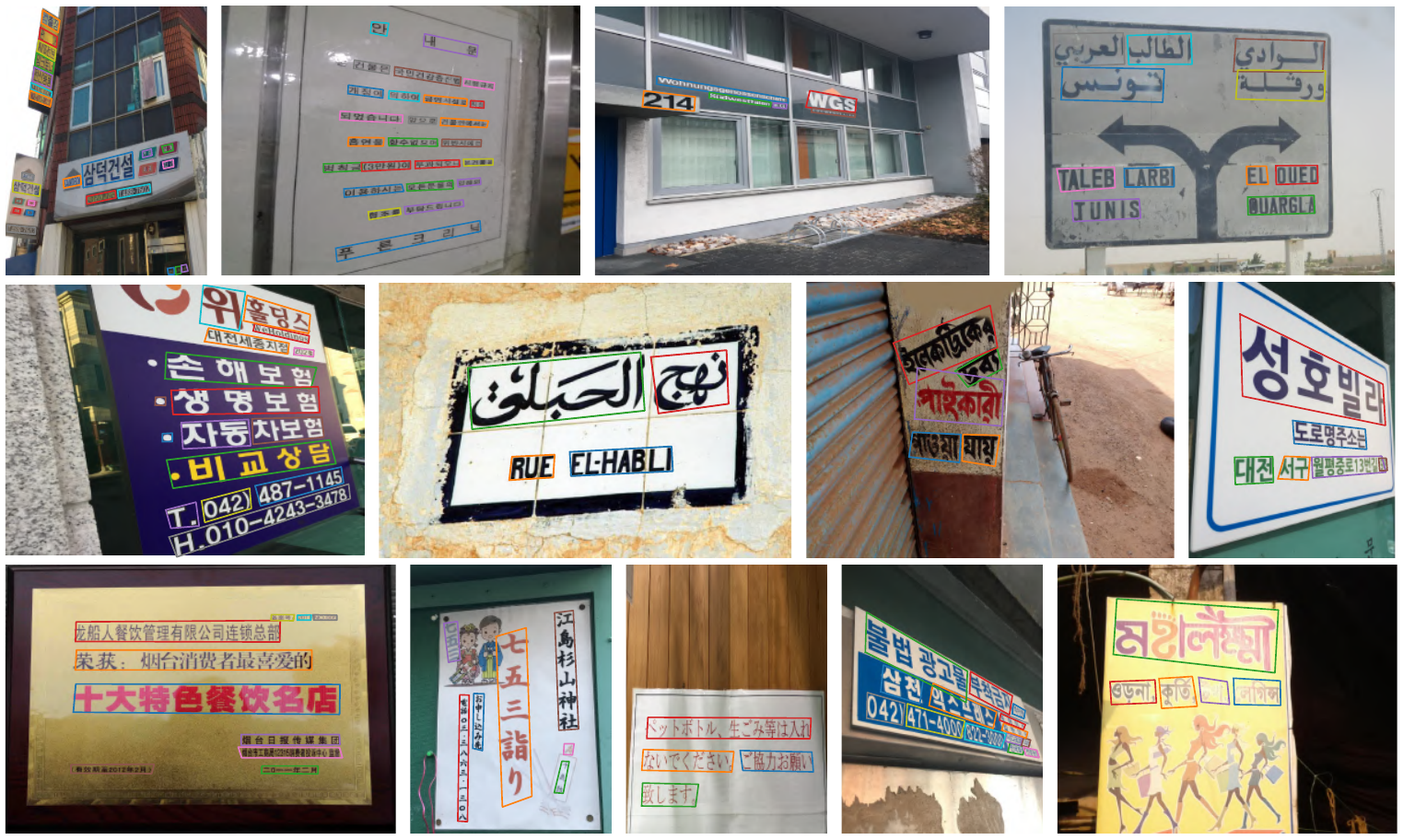}\label{fig:qualitative_detection_results}}
    \subfigure[End-to-end results on ReCTS.]{\includegraphics[width=0.95\textwidth]{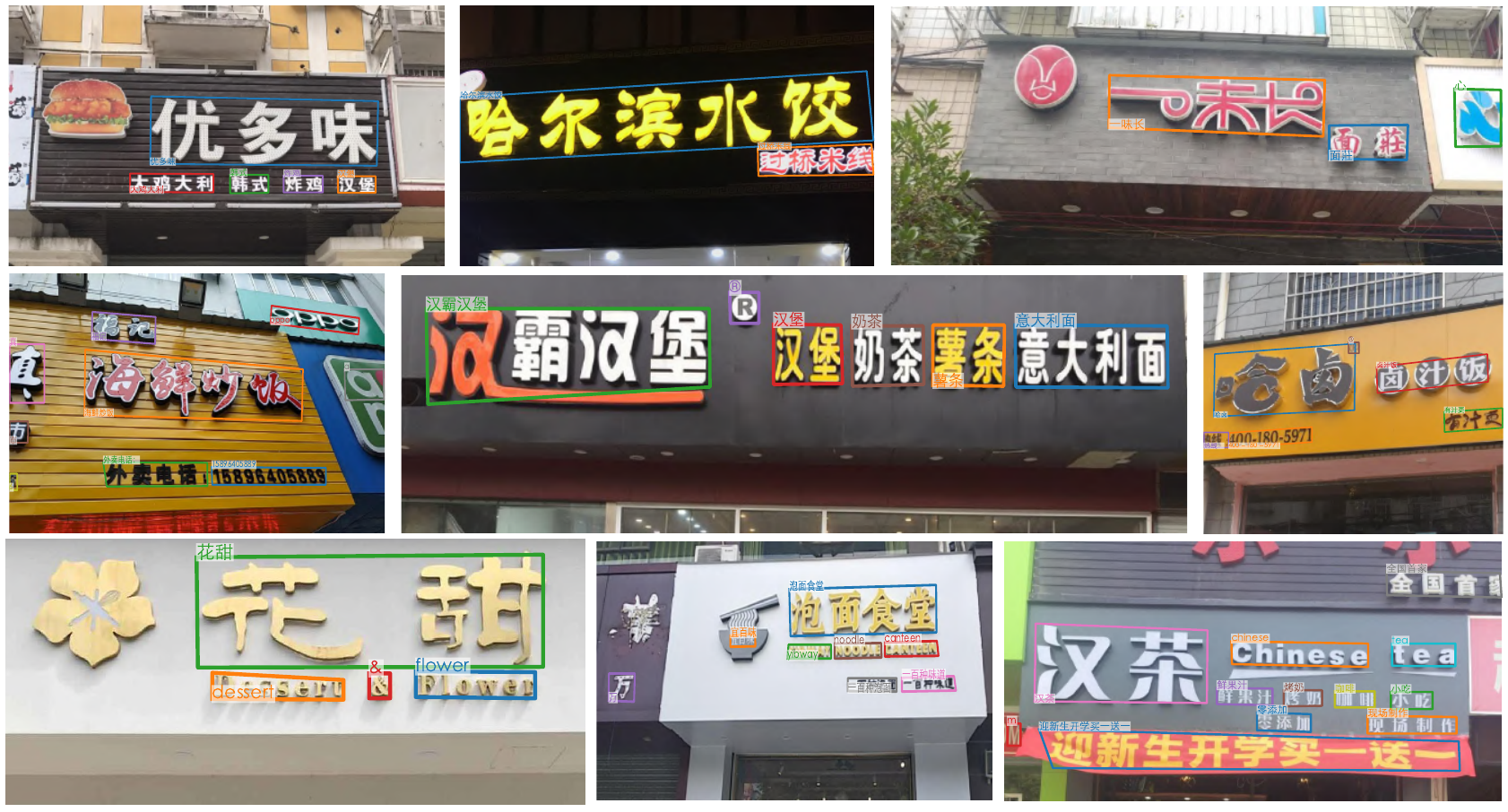}\label{fig:qualitative_end2end_results}}
    \caption{Visualization of the qualitative results outputted by the proposed approach.}
    \label{fig:my_label}
\end{figure*}

  {\bf ReCTS End-to-End.} One of the main goals of scene text detection is to recognize a text instance \cite{xie2019convolutional} that is highly related to the performance of the detection system. To validate the effectiveness and robustness of our detection method, we build  a recognition system that incorporate several state-of-the-art methods. 
  Typically, the recognition performance is highly relevant to the quality of the detected boxes. To reveal the precision of our detection, we construct an end-to-end recognition system to demonstrate how our method benefits recognition models. We first crop the images using detected boxes and fed them into four popular recognition models, including decouple attention network \cite{wang2020Decoupled}, convolutional recurrent neural network \cite{shi2017end}, network of show, attend, read \cite{li2019show}, and transformer-based networks \cite{wang2019Simple}. The four models 
  are
  trained on real samples and 600,000 extra synthetic samples following their default settings for training. The real samples are provided by the official training set, whereas the synthetic samples are synthesized using a render engine \cite{jaderberg2016reading} and the corpus of the official training set. All images are resized to a specific required height for each recognition model while maintaining the aspect ratio of the original image. In a data batch, all images are padded with white to the maximum width of the images. During the inference stage, we choose the prediction having the highest confidence as the final ensemble result. Both quantitative and qualitative results are presented in Table \ref{tab:rects} and Figure \ref{fig:qualitative_end2end_results}, respectively.

\begin{figure*}
    \centering
    \includegraphics[width=14.53cm]{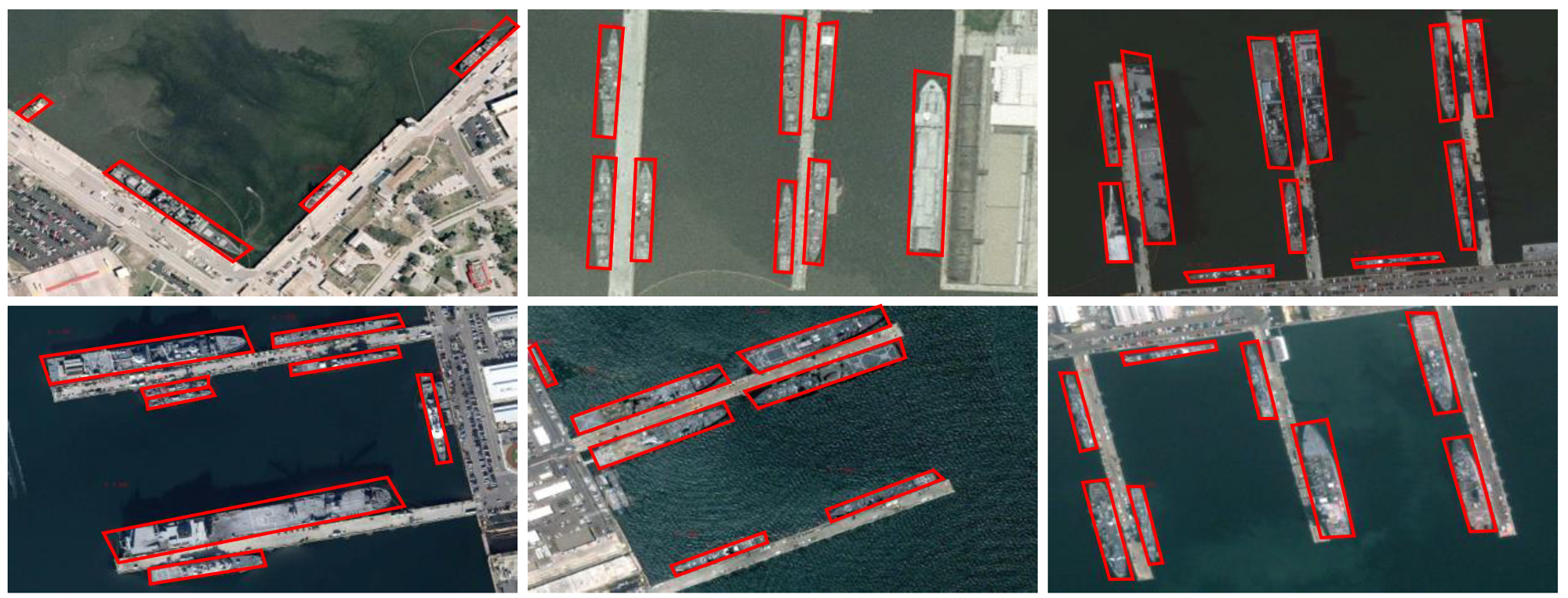}
    \caption{Qualitative detection results on the HRSC2016 dataset.}
    \label{fig:hrsc_vis}
\end{figure*}

  {\bf HRSC2016.} To demonstrate the generalization ability of our method, we further evaluate its performance on the Level-1 task of the HRSC2016 dataset \cite{liu2017rotated} to demonstrate multi-directional object detection. The ship instances in this dataset are presented in various orientations, and the annotating bounding boxes are based on rotated rectangles. There were 436, 181, and 444 images for training, validating, and testing, respectively. Only the training and validation sets are used for training. The evaluation metric is the same as in \cite{liu2019omnidirectional,karatzas2015icdar}. The result is shown in Table \ref{tab:hrsc}, showing a significant improvement over the TIoU-Hmean \cite{liu2019tightness}. It also demonstrates the robustness of our method. Qualitative examples of the detection results are shown in Figure \ref{fig:hrsc_vis}.

\begin{table}[!t]
\centering
\small
\begin{tabular}{c|ccccc}
  \hline
  Algorithms & R (\%) & P (\%) & H (\%) & TIoU-H (\%) & mAP \\
  \hline
  \cite{girshick2015fast,liao2018rotation}&-&-&-&- & 55.7\\
  \cite{girshick2015fast,liao2018rotation}&-&-&-&- & 69.6\\
  \cite{girshick2015fast,liao2018rotation}&-&-&-&- & 75.7\\
  \cite{liao2018rotation}&-&-&-&- & 84.3 \\
  \hline
  \hline
 Liu et al. \cite{liu2019omnidirectional}& 94.8 & 46.0 & 61.96 & 51.1 & 93.7 \\
  \hline
  Ours & 94.1 & 83.8 & \bf 88.65 & \bf 73.3 & 89.22 \\
  \hline
  Ours (low cf) & 95.7 & 54.2 & 69.2 & 57.5 & \bf94.8 \\
  \hline
\end{tabular}
\caption{Experimental results for HRSC\_2016 dataset. cf: confidence threshold, which is set to 0.01 in the last line.}
\label{tab:hrsc}
\end{table}

\section{Conclusion}

In this paper, we have addressed \multioriented scene text detection using an effective \Ours method. Using discretization methodology, \Ours, can solve the inconsistent labeling issue by discretizing the point-wise prediction into orderless key edges. To decode accurate vertex positions, we have proposed a simple but effective MTL method to reconstruct the quadrilateral bounding box. Benefiting from \Ours, we improve the reliability of the confidence of the bounding box and adopted more effective post-processing methods to improve performance.

Additionally, based on our method, we have conducted thorough ablation studies on six training components, including data arrangement, pre-processing, backbone, proposal generation, prediction head, and post-processing, to explore the potential upper limit of our method. By %
combining
effective modules, we have achieved state-of-the-art results on various benchmarks and won the first place in the recent ICDAR 2019 Robust Reading Challenge on Reading Chinese Text on Signboards. Moreover, using a recognition model, we perform the best in the end-to-end detection and recognition task, verifying that our method is conducive to current recognition methods. To test the generalization ability, we have conducted an experiment on an oriented general object dataset HRSC2016; the results verify that our method can significantly outperform recent state-of-the-art methods.

\small 

\bibliographystyle{ieeetr}

\bibliography{IJCAItoIJCV}   %

\end{document}